\pdfoutput=1
\documentclass[11pt]{article}

\usepackage{acl}

\usepackage{times}
\usepackage{latexsym}

\usepackage[utf8]{inputenc} % allow utf-8 input
\usepackage[T1]{fontenc}    % use 8-bit T1 fonts
\usepackage{url}            % simple URL typesetting
\usepackage{booktabs}       % professional-quality tables
\usepackage{amsfonts}       % blackboard math symbols
\usepackage{nicefrac}       % compact symbols for 1/2, etc.
\usepackage{microtype}      % microtypography
\usepackage{xcolor}         % colors
\usepackage{booktabs}       % professional-quality tables
\usepackage{amsfonts}       % blackboard math symbols
\usepackage{nicefrac}       % compact symbols for 1/2, etc.
\usepackage{microtype}      % microtypography
\usepackage{amsbsy,amsthm,amssymb,color,xcolor}

\usepackage{multicol}
\usepackage{multirow}
\usepackage{caption}
\usepackage{subcaption}
\usepackage{wrapfig}
\usepackage{xspace}
\usepackage{paralist}
\usepackage{makecell}
\usepackage{color}
\usepackage{dsfont}
\usepackage{makecell}
\usepackage{graphicx}
\usepackage{amsmath,amsfonts,bm}

\def\eqref#1{equation~\ref{#1}}
\def\1{\bm{1}}

\def\vx{{\bm{x}}}
\def\vy{{\bm{y}}}

\DeclareMathAlphabet{\mathsfit}{\encodingdefault}{\sfdefault}{m}{sl}
\SetMathAlphabet{\mathsfit}{bold}{\encodingdefault}{\sfdefault}{bx}{n}

\def\gP{{\mathcal{P}}}

\newcommand{\E}{\mathbb{E}}
\newcommand{\Ls}{\mathcal{L}}

\newcommand{\KL}{D_{\mathrm{KL}}}

\DeclareMathOperator*{\argmax}{arg\,max}

\usepackage{algorithm}
\usepackage{algorithmic}
\usepackage{lipsum}

\usepackage{wrapfig}
\newcommand{\tempneg}{-}
\newlength\myindent
\newcommand{\method}{\textsc{DiffGlat}\xspace}

\newcommand{\RN}[1]{%
  \textup{\uppercase\expandafter{\romannumeral#1}}%
}

\title{Diffusion Glancing Transformer for Parallel Sequence-to-Sequence Learning}

\author{Lihua Qian$^{1}$ \enskip Mingxuan Wang$^{1}$ \enskip Yang Liu$^{2}$ \enskip Hao Zhou$^{2}$\\
$^1$ ByteDance \enskip $^2$Institute for AI Industry Research (AIR), Tsinghua University\\
\texttt{\{qianlihua,wangmingxuan.89\}@bytedance.com}\\
\texttt{{liuyang2011@tsinghua.edu.cn}} \\
\texttt{{zhouhao@air.tsinghua.edu.cn}} \\
}

\begin{document}
\maketitle
\begin{abstract}
Previously, non-autoregressive models were widely perceived as being superior in generation efficiency but inferior in generation quality due to the difficulties of modeling multiple target modalities.
To enhance the multi-modality modeling ability, we propose the diffusion glancing transformer~(\method), which employs a modality diffusion process and residual glancing sampling.
The modality diffusion process is a discrete process that interpolates the multi-modal distribution along the decoding steps, and the residual glancing sampling approach guides the model to continuously learn the remaining modalities across the layers. 
Experimental results on various machine translation and text generation benchmarks demonstrate that \method achieves better generation accuracy while maintaining fast decoding speed compared with both autoregressive and non-autoregressive models. 
%\zhouh{leave it until last minitues}
% For sequence \textit{generation}, both autoregressive models and non-autoregressive models have been developed in recent years. Autoregressive models can achieve high generation quality, but the sequential decoding scheme causes slow decoding speed. Non-autoregressive models accelerate the inference speed with parallel decoding, while their generation quality still needs to be improved due to the difficulty of modeling multi-modalities in data. 
% To address the multi-modality issue, we propose \method, a non-autoregressive model featured with a modality diffusion process and residual glancing training. The modality diffusion process decomposes the modalities and reduces the modalities to learn for each transition. And the residual glancing sampling further smooths the modality learning procedures.
% Experiments demonstrate that, without using knowledge distillation data, \method can achieve superior performance in both decoding efficiency and accuracy compared with the autoregressive Transformer.

\end{abstract}

\section{Introduction}
\label{sec:intro}

The Transformer~\citep{transformer2017vaswani} has been the most widely used architecture in sequence generation, outperforming its counterparts in (almost) all downstream tasks, such as machine translation and question answering~\cite{gpt3,openai2023gpt4}.
The typical transformer architecture adopts the autoregressive decoding strategy~(AR), generating tokens in a predefined order, i.e., left to right. 
Recently, non-autoregressive generation models~(NAR) attract increasing attention for their fast generation speed, which are \textit{considered} to sacrifice generation quality by generating the outputs simultaneously instead of sequentially.

However, we argue that NAR has advantages compared to AR \textit{beyond generation efficiency} for the following reasons: 1) the parallel generation enables NAR to remove the \textit{inductive bias} of the \textit{predefined generation order}, thereby liberating the potential of applications, such as molecules or proteins that have no well-defined orders; 
% The autoregressive generation could hurt the performance for the tasks where organized thinking are required~\citep{bubeck2023sparks}, or targets that have no well-defined orders, such as molecules/proteins\footnote{The recycling module used in AlphaFold2~\citep{jumper2021highly}}; 
% 2) the unidirectional generation paradigm can hinder the model from organized thinking.~\citep{bubeck2023sparks}
2) furthermore, NAR can utilize \textit{bi-directional context} for sequence modeling and generation, whereas AR could only exploit context from one direction.
% Specifically, NAR could outperform AR not only in efficiency but also in accuracy, which has potentials to replace AR as the mainstream network architecture in sequence to sequence learning for following reasons:
% \begin{itemize} 
%     % \item As is well known, NAR generates outputs in parallel instead of one after another, it could fully utilize the parallelism of GPU devices and lead to fast generation speed.
%     \item Furthermore, the characteristic of parallel generation enables NAR to remove the bias of predefined order in generation, which liberates the generation process. Specifically, except sentence, many generation targets do not have a well-defined order, and generating them in an auto-regressive way could introduce permutation bias\footnote{xxx}\zhouh{add footnote to explain this}, thus hurting the generation performance (such as molecule/protein).\zhouh{citation} 
%     \item Finally, NAR by design can utilize bidirectional context for sequence modeling and generation. On the contrary, AR could only exploit context from one direction for generating the target sequence, which is insufficient, especially when the context from the other direction is crucial for making prediction.
% \end{itemize}
\begin{figure}[t]
\centering
\includegraphics[width=0.8\linewidth]{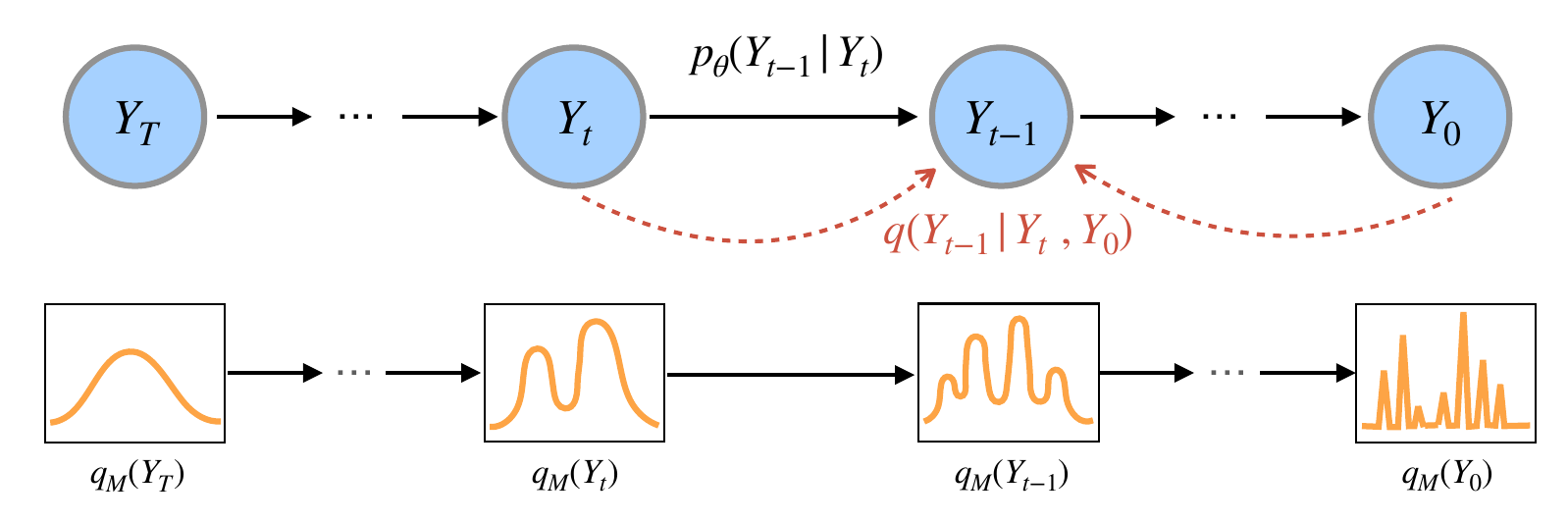}
\caption{Our modality diffusion process. $q(\bm{y}_t)$ represents the marginal distribution of $\bm{y}_t$, which visualizes the modality distribution. The number of modalities increases as the timestep $t$ decreases.}
\label{fig:MDP}
\vspace{-1em}
\end{figure}

Although being promising, current state-of-the-art NAR models still fall behind the AR counterpart in terms of generation quality. 
The main drawback of NAR lies in the difficulty of modeling multi-modal distributions~\citep{nat2018gu}, i.e., one input has multiple valid outputs.
% Recent progress makes NAR approach AR in accuracy.
% However, NAT suffers from a tough learning problem---the multi-modality issue~\citep{nat2018gu}---that not encountered by the auto-regressive Transformer~(AT), which makes NAT's performance fall behind AT.
For example, a source sentence could be translated into multiple different target sentences. As the NAR model decodes all the tokens in parallel, it may mix tokens from different translation targets. In contrast, the AR model can output consistency tokens more easily as every prediction is conditioned on previous ones.
% This problem is caused by the parallel generation nature of NAR, which generates tokens \textit{independently} and does not \textit{explicitly} propagate information about the final predictions~(as in AR) among different time steps of the sentence.
Recent progress of NAR has significantly improved the multi-modal learning ability of parallel generation, including knowledge distillation~\citep{seqkd2016kim}, iterative decoding~\citep{iterativerefinement2018lee}, latent variable modeling~\citep{flowseq2019ma} and revised learning objectives~\citep{ctc2018libovicky,axe2020ghazvininejad,oaxe2021du,dag2022huang}. However, these NAR models still hardly outperform the AR baseline consistently.
% Fortunately, the problem has been greatly eliminated with recent progress in NAR.

% Recent work of NAR greatly pushes forward the multi-modal modeling ability of NAR in three perspectives:
% 1) Introducing iterative decoding and latent variables to explicitly captures the dependencies with multiple decoding iterations or partial information in the latent variables~\citep{iterativerefinement2018lee, flowseq2019ma}.
% 2) Additionally, gradual training strategies, such as glancing training~\citep{glat2021qian}, significantly boosts the multi-modality learning capacity for non-iterative NAR via adaptive task learning.
% 3) Finally, learning objectives like CTC and DAG~\citep{ctc2018libovicky, dag2022huang} improve the NAR multi-modal learning by target alignment, which enables the NAR models to avoid the mismatch between the target and the model output in training.
% To this end, compared with the strong AR baseline, latest NAR models like DA-Transformer can achieve comparable generation quality without knowledge distillation.
% Notably, NAR machine translation systems have been adopted to win the first place in WMT German$\rightarrow$English 2021 news translation shared task~\citep{qian-etal-2021-volctrans}, which has been deployed in products for large scale users.

In this paper, we propose \method, which shows that NAR can outperform AR in both \textit{efficiency} and \textit{accuracy}, without requiring knowledge distillation from AR.
Generally \method is designed within the denoising diffusion implicit model framework~\cite{song2021denoising}. 

First, to reduce the difficulty of learning multi-modalities for NAR, \method defines a discrete \textit{modality diffusion process} that \textit{smoothly} decomposes the modality learning in the data across many diffusion transition steps. 
With modality decomposition, each diffusion transition only includes a scheduled number of modalities, which makes the modality learning much easier for NAR. And our preliminary experiments confirm the effectiveness of modality decomposition~(See Section \ref{sec:preliminary}).
Note that each diffusion transition is learned by several adjacent NAR layers, and thus \method can stack a sufficient number of layers to model complex multi-modal distributions.

Besides, we proposes a \textit{residual glancing sampling}  technique, which adaptively adjusts the learning difficulty of each diffusion transition in a layer-wise and residual manner.
%Our approach improves NAR layers' modeling capacity of multi-modal distributions and achieves satisfactory generation results with few denoising transitions and NAR layers.

% Different from directly applying the diffusion method in~\citet{diffusion2020}, our proposed modality diffusion process, which works with the residual glancing together, achieving superior generation performance with high decoding efficiency.

Experiments demonstrate that the proposed \method significantly improves the quality of NAR generation on standard benchmarks. Using only 1 decoding iteration, \method can outperform all NAR baselines, including iterative ones. With 3 iterative decoding steps, \method beats the strong AR baseline with a moderate margin~(+0.47 BLEU). Comparisons between \method and AR Transformer on 10 standard machine translation benchmarks, with both Transformer base and big settings, consistently verify the effectiveness of \method.
Additionally, we also find that \method can slightly outperform the AR baseline on image captioning and paraphrasing tasks. These results show the extensive potential of \method in other generation tasks.

\section{Preliminary}
\label{sec:background}

% Given the input sequence $\bm{X}={x_1, x_2, ..., x_m}$ and target sequence $\bm{Y}={y_1, y_2, ..., y_n}$, the NAR model factorizes the joint probability $P(\bm{Y}|\bm{X})$ with the conditional independence assumption:
% \begin{equation}
%  p(\bm{Y}|\bm{X};\theta) = \prod \limits_{i=1}^n p(y_i|\bm{X};\theta),\\
% \end{equation}
% where $\theta$ is the parameter of the model, and each token $y_i$ in $\bm{Y}$ is independent from other target tokens. 

% Intuitively, NAR imposes a conditionally independent factorization in the generation process, which generates sentences in a parallel way and requires the model to capture the modalities for the combination of all the tokens simultaneously. 
% It can be quite challenging while the target multi-modal distribution is complex. when the number of modalities is large. In contrast, the AR model factorizes the joint probability in an autoregressive way:
% \begin{equation}
%     p(\bm{Y}|\bm{X};\theta) = \prod \limits_{i=1}^n p(y_i|y_{< i},\bm{X};\theta),\\
% \end{equation}
% By conditioning on preceding tokens $y_{< i}$, the autoregressive factorization divides the modality learning of the sequence into multiple learning steps for tokens. Since AR only predict one token in each decoding step, the modality to be learned for each step is considerably less than that of the combination of all the tokens.
Given the input sequence $\bm{x}={x^1, x^2, ..., x^m}$ and the target sequence $\bm{y}={y^1, y^2, ..., y^n}$, the NAR model factorizes the joint probability $P(\bm{y}|\bm{x})$ with the conditional independence assumption:
\begin{equation}
 p(\bm{y}|\bm{x};\theta) = \prod \limits_{i=1}^n p(y^i|\bm{x};\theta),\\
\end{equation}
where $\theta$ is the parameter of the model, and each token $y^i$ in $\bm{y}$ is independent from other target tokens. The conditionally independent factorization demands the model to capture the modalities for the combination of all the tokens in one step, which can be difficult when the number of modalities is large. In contrast, the AR model factorizes the joint probability in an autoregressive way:
\begin{equation}
    p(\bm{y}|\bm{x};\theta) = \prod \limits_{i=1}^n p(y^i|y^{< i},\bm{x};\theta),\\
\end{equation}
By conditioning on preceding tokens $y^{< i}$, the autoregressive factorization divides the modality learning of the sequence into multiple steps where each step learns to predict one token. Besides the autoregressive factorization, the denoising diffusion models smooth the source-target transformation by interpolating intermediate distributions between the source and the target.
%Since AR only predict one token in each decoding step, the modality to be learned for each step is considerably less than that of the combination of all the tokens.

Inspired by the learning decomposition in AR and diffusion models, we employ training procedures as in diffusion models to decompose the complex modality learning of NAR into several easier diffusion transition steps. Since the process of adding Gaussian noise is not designed for tackling the multi-modality problem, we also explore new diffusion processes to address the problem. A preliminary study demonstrates that a learning process with a gradually growing number of modalities is beneficial for learning modalities in NAR.

% \subsection{From DDPM to DDIM}
% Motivated by the intuition of probabilistic diffusion model, we propose \method that decomposes the unaffordable modality learning of NAR into many diffusion transition steps. 
% However, we find that the typical \textit{denoising diffusion probabilistic model}~(DDPM) realizes diffusion modeling by adding Gaussian noise, which is inefficient and is not necessarily the best choice in NAR modeling. 
% Thus we choose the \textit{denoising diffusion implicit model}~(DDIM) as our probabilistic modeling framework, and define specific diffusion and generative processes to implement the proposed \method. 
% The main difference between DDPM and DDIM lies in the different factorization of the posterior $q(\bm{y}_{1:T}|\bm{y}_0)$, the former one is Markovian and the latter one is non-Markovian. %Please refer to Appendix~\ref{sec:diffusion_detail} for more details.
\subsection{Denoising Diffusion Models}
With a series of latent variables $\bm{y}_1\bm{y}_2...\bm{y}_T$,
the diffusion process can be characterized by the posterior $q(\bm{y}_{1:T}|\bm{y}_0)$. This process guides the generative process $p(\bm{y}_{0:T}):= p_\theta(\bm{y}_T)\prod_{t=1}^T p_\theta(\bm{y}_{t-1}|\bm{y}_{t})$ in diffusion models~\citep{thermo2015, diffusion2020} to fit the data $\vy_0$ step by step. Depending on the Markov property of the process defined by $q(\bm{y}_{1:T}|\bm{y}_0)$, the diffusion processes can be divided into Markovian and non-Markovian ones.

\paragraph{Markovian Diffusion Process}
The Markovian diffusion processes are employed in many previous work for diffusion models~\citep{thermo2015,diffusion2020}.
In these work, the forward process is a Markov chain where the posterior for $\bm{y}_t$ can be determined by conditioning on $\bm{y}_{t-1}$:
\begin{equation}
    q(\bm{y}_{1:T}|\bm{y}_0):= \prod_{t=1}^{T}q(\bm{y}_t|\bm{y}_{t-1})
\end{equation}
For example, \citet{diffusion2020} propose a denoising diffusion probabilistic model~(DDPM), which adds Gaussian noise with increasing variances $\beta_{1:T} \in (0,1]^{T}$, and the forward transition probability is defined by: $q(\bm{y}_t|\bm{y}_{t-1}):=\mathcal{N}(\bm{y}_{t};\sqrt{1-\beta_t}\bm{y}_{t-1},\beta_t \mathbf{I})$. The posterior $q(\bm{y}_{t-1}|\bm{y}_t,\bm{y}_0)$ of DDPM can be computed in closed form, and the detailed derivation can be found in ~\citep{diffusion2020}.
\paragraph{Non-Markovian Diffusion Process}
Besides the Markovian process, the posterior $q(\bm{y}_{1:T}|\bm{y}_0)$ can also be modeled as a non-Markovian process~\citep{song2021denoising}:
\begin{equation}
q(\bm{y}_{1:T}|\bm{y}_0):=q(\bm{y}_T|\bm{y}_0)\prod_{t=2}^{T}q(\bm{y}_{t-1}|\bm{y}_t,\bm{y}_0),
\end{equation}
where each decomposition term depends on $\bm{y}_0$.
%Different from the Markovian process, we can obtain $q(\bm{y}_{t-1}|\bm{y}_t,\bm{y}_0)$ from the non-Markovian factorization directly. And the forward transition probability can be computed using the Bayes' rule.
% \begin{equation}
% q(\bm{y}_t|\bm{y}_{t-1},\bm{y}_0)=\frac{q(\bm{y}_{t-1}|\bm{y}_t,\bm{y}_0)q(\bm{y}_t|\bm{y}_0)}{q(\bm{y}_{t-1}|\bm{y}_0)}
% \end{equation}
%The forward transition probability is actually not used if the generative model is trained with Eq.~\ref{eqn:vlb}
According to~\citep{song2021denoising}, non-Markovian processes are less stochastic than the Markovian process defined in ~\citet{diffusion2020}, leading to more deterministic and efficient generative models. Thus, we propose to use a non-Markovian process to decompose the modalities in data.

\begin{table}[t]
\begin{center}
\begin{small}
\resizebox{\linewidth}{!}{
\begin{tabular}{c|lll|l}
\toprule
 Source & \multicolumn{3}{c|}{Modality} & \multicolumn{1}{c}{Target}\\
\midrule
\multirow{6}{*}{$\mathtt{52\ \ 1\ \ 937\ \ 1234}$} & $\RN{1}$ &$y^i=x^i+5000$, $i$ is odd; &$y^i=x^i+10000$, $i$ is even & $\mathtt{5052\ \ 10001\ \ 5937\ \ 11234}$ \\
% &  &$y^i=x^i+10000$, $i$ is even &  \\
\cmidrule{2-5}
& $\RN{2}$ &$y^i=x^i+10000$, $i$ is odd; &$y^i=x^i+5000$, $i$ is even & $\mathtt{10052\ \ 5001\ \ 10937\ \ 6234}$\\
% &  &$y^i=x^i+5000$, $i$ is even &  \\
\cmidrule{2-5}
& $\RN{3}$ &$y^i=x^i+15000$, $i$ is odd; &$y^i=x^i+20000 $, $i$ is even & $\mathtt{15052\ \ 20001\ \ 15937\ \ 21234}$\\
% &  & $y^i=x^i+20000 $, $i$ is even &  \\
\cmidrule{2-5}
& $\RN{4}$ &$y^i=x^i+20000$, $i$ is odd; &$y^i=x^i+15000 $, $i$ is even& $\mathtt{20052\ \ 15001\ \ 20937\ \ 16234}$\\
% &  & $y^i=x^i+15000 $, $i$ is even &  \\
\bottomrule
\end{tabular}
}
\end{small}

\caption{Illustration of the synthetic data.}
%For instance, the first modality adds 5000 to the numbers on the odd digits and adds 10000 to the numbers on the even digits, thus the first number in the target is $52+5000=5052$ and the second number is $1+10000=10001$.}
\label{tab:synthetic_data}
\vspace{-1em}
\end{center}
\end{table}

\begin{table}[t]
% \begin{minipage}{0.5\linewidth}
% \begin{table}[t]
\centering
\begin{small}
\resizebox{0.9\linewidth}{!}{
\begin{tabular}{l|ccccc}
\toprule
\multirow{2}{*}{Model}& \multirow{2}{*}{DATA} &\multirow{2}{*}{AR} & \multirow{2}{*}{NAR} & \multicolumn{2}{c}{NAR w/ PML}\\
& & & &Middle&Last\\
\midrule
Token Acc. & 100 &100.0 & 56.6 & 98.8 & 97.7\\
Seq Acc. & 100& 100.0 & 0.0 & 93.4 & 88.6\\
% \midrule
% \makecell[c]{Modality\\ Distribution} & \makecell[c]{[24.9\% 25.8\%\\ 24.1\% 25.2\%]} &  \makecell[c]{[20.1\% 27.8\% \\ 32.0\% 20.1\%]} & \makecell[c]{[29.6\% 22.4\%\\  27.7\% 20.3\% ]} & \makecell[c]{[0 50.0\%\\  0 50.0\%]} & \makecell[c]{[25.5\% 23.8\%\\  28.3\% 22.4\%]}\\
\bottomrule
% Sequence Accuracy &
%  & Number & Sequence\\% & Modality Distribution\\
%  & Accuracy& Accuracy\\
% \midrule
% Data & 100.0 & 100.0 \\%& [24.9\%, 25.8\%, 24.1\%, 25.2\%]\\
% AR & 100.0 & 100.0 \\%& [20.1\%,27.8\%,32.0\%,20.1\%]\\
% NAR & 56.6 & 0.0 \\%& [29.6\%, 22.4\%, 27.7\%, 20.3\% ]\\
% NAR+GM~(Layer2) & 98.8 & 93.4 \\%& [0, 50.0\%, 0, 50.0\%]\\
% NAR+GM~(Layer4) & 97.7 & 88.6 \\%& [25.5\%, 23.8\%, 28.3\%, 22.4\%]\\
% \bottomrule
\end{tabular}}
\end{small}
\caption{Results of synthetic experiments.}
%PML represents the model is trained with progressive modality learning.}
%Layer2 is the middle layer of decoder, and Layer4 is the last layer that outputs the final representations. }
\label{tab:synthetic_experiment}
\vspace{-1em}
\end{table}
\begin{figure}[t]
% \end{table}
% \end{minipage}\hfill
% \begin{minipage}{0.5\linewidth}
% \begin{figure}[t!]
\centering
\includegraphics[width=0.62\linewidth]{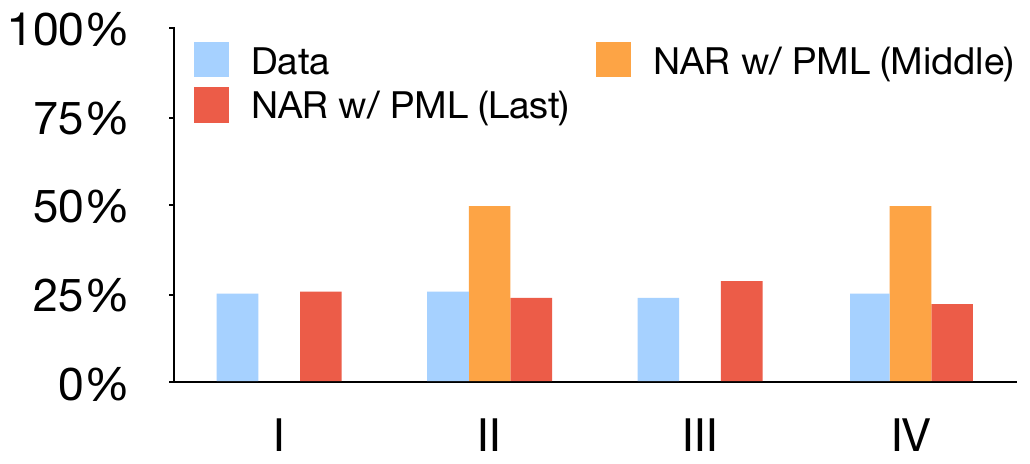}
\vspace{-0.5em}
\captionof{figure}{The output modality distributions.}
\label{fig:modality_distribution}
% \end{minipage}
\vspace{-1.2em}
\end{figure}
\subsection{Proof of Concept}
\label{sec:preliminary}
In addition,  we also include a synthetic experiment to demonstrate that learning with growing number of modalities across layers can effectively benefit the performance of NAR.
% To verify the effectiveness of progressive modality learning~(PML), we devise synthetic experiments where the number of modalities can be explicitly manipulated.
%The experimental result shows that the NAR model can capture modalities more accurately by gradually increasing the number of modalities to learn.
Specifically, we create a synthetic dataset with the 4 modalities shown in Table~\ref{tab:synthetic_data}, where each source has only one target but may come from one of four different modalities. 
% Since we can control the modalities in such synthetic data, we train a NAR model to first learn 2 modalities, and then all the modalities.
We simulate the modality diffusion process by using the progressive modality learning~(PML) inside NAR layers. 
In PML, we train the middle layers of NAR with only 2 modalities by transforming the target of modality $\RN{1}$ to modality $\RN{2}$ and the target of modality of $\RN{3}$ to modality $\RN{4}$. 
And we train the last decoder layer by the targets with 4 modalities (See Appendix~\ref{sec:synthetic} for more details). 

The results in Table~\ref{tab:synthetic_experiment} show that the vanilla NAR model fails to output correct sequences, and both the middle and last layer of the NAR trained with PML achieve significant accuracy gains. 
From the modality distribution plot in Figure.~\ref{fig:modality_distribution}, we can find that the NAR trained with PML learns only 2 modalities in the middle layer, and captures all the modalities in the last layer.

\section{The Proposed Diffusion-GLAT}

\label{sec:method}
% \begin{itemize}
%     \item Figure for illustrating the diffusion training %decomposition of the entire dependency learning task
%     \item Method
%         \begin{enumerate}
%         \item the diffusion process defined in our work $q(\bm{y}_{t-1}|\bm{y}_t, \bm{y}_0)$
%         \item glancing training (parameterization)
%         \end{enumerate}
%     \item mode coverage
%     \item objective
%         \begin{enumerate}
%         \item loss (cross entropy)
%         \item ctc and dag (summary)
%     \end{enumerate}
%     \item inference
% \end{itemize}

In this section, we will introduce \method, which enables parallel sequence to sequence learning in a denoising diffusion implicit model~(DDIM, \citealp{song2021denoising}) framework.
The primary goals of designing \method are: a) decomposing the modality learning to reduce the training difficulty of NAR and b) achieving high generation quality with few denoising transitions to keep fast decoding. 
%improving the modalities captured by each layer to cover modalityies as more as possible in finite Transformer layers. 
%To the achieve the , we 

% Generally, \method is designed in a probabilistic diffusion framework, which consists of a \textit{generative process} $p_{\theta}(\vy_{0:T}|\vx)$ and a \textit{diffusion process} $q(\vy_{1:T}|\vy_{0},\vx)$, where $\vx=\{x^1,x^2,...,x^N\}$ is the source sequence, $\vy_{0}=\{y^1,y^2,...,y^M\}$ is the target sequence, $N, M$ are the corresponding sequence lengths, and $\vy_1, \vy_2, \ldots, \vy_T$ is a series of latent variables.

% During inference, the generated sequences are obtained by maximizing  $p_{\theta}(\vy_0|\vx) = \int p_\theta(\vy_{0:T}|\vx) \mathrm{d} \vy_{1:T}$ with the generative process, where  $p_\theta(\vy_{0:T}|\vx):=p(\vy_T|\vx)\prod_{t=1}^T p_\theta(\vy_t|\vy_{t-1},\vx)$.\qlh{need fixing}
%and $p_\theta$ are parameterized by NAR Transformers.

In denoising diffusion models, the parameters $\theta$ are trained to fit the data distribution $q(\vy_0)$ by maximizing a variational lower bound~\citep{diffusion2020}:
\begin{align}
\label{eqn:method_vlb}
    &\Ls_\text{VLB} = \E_{q(\vy_{0}|\vx)}\Big[-\log p_\theta(\vy_{0}||\vx)\Big] \leq \\
    &\E_{q(\vy_{0:T}|\vx)}\Big[\KL \big( q(\vy_{1:T} | \vy_0,\vx)||p_\theta(\vy_{0:T}|\vx)\big)\Big] \nonumber
\end{align}

% However, adding Gaussian noises as in the DDPM~\citep{diffusion2020} does not match the discrete nature of data like text. 
To achieve the two goals described in the beginning, we define a discrete diffusion process $q^\text{MDP}$ that adds the modalities in the data gradually as the diffusion step $t$ decreases. Additionally, the training process is equipped with the residual glancing sampling techniques $p_\theta^\text{RGS}$, which further boosts the modality learning ability.
Therefore, we can rewrite the training objective in Eq.~\ref{eqn:method_vlb} as:
% \begin{equation}
%     L = \sum_{t> 1}\underbrace{D_{\text{KL}}(q(\bm{y}_{t-1}|\bm{y}_t,\bm{y}_0,\bm{x})||p_\theta (\bm{y}_{t-1}|\bm{y}_t,\bm{y}_{t-1}^\text{gla},\bm{x}))}_{L_{t-1}^\text{gla}} -\mathbb{E}_{q(\bm{y}_1|\bm{y}_0,\bm{x})}\log p_\theta(\bm{y}_0|\bm{y}_1, \bm{y}_0^\text{gla},\bm{x})
% \end{equation}
\begin{align}
    \label{eqn:our_vlb}
    \Ls_\text{VLB} = & \mathbb{E}_{q}D_\text{KL}(q(\bm{y}_{1:T}|\bm{y}_0,\bm{x})|p_\theta(\bm{y}_{0:T}|\bm{x})) \\
    = & \mathbb{E}_{q}D_\text{KL}(q^\text{MDP}(\bm{y}_{1:T}|\bm{y}_0,\bm{x})|| p_\theta^\text{RGS}(\bm{y}_{0:T}|\bm{x})) \nonumber
\end{align}

The modality diffusion process $q^\text{MDP}$ gradually interpolates modalities among intermediate layers, which adaptively schedules sequence modalities across the NAR Transformer layers.
And the residual glancing sampling~(RGS) samples the target tokens that are not correctly captured across neural layers to enhance the modality learning process.

\subsection{Modality Diffusion Process: $q^\text{MDP}$}
\label{sec:modality_diffusion}
In order to reduce the difficulty of learning modalities for NAR, we introduce a discrete diffusion process that distributes modalities to multiple transitions, which is illustrated in Figure~\ref{fig:MDP}.
As shown in Section~\ref{sec:preliminary}, a reasonable denoising process for learning modalities is adding the modalities in $\vy_0$ gradually. 
But the modalities in the real world data cannot be explicitly manipulated, thus we need to divide the modalities in $\vy_0$ implicitly. Previous work reveals that the trained NAR models could capture part of the modalities in the data~\citep{kdnat2020zhou}. Therefore, we leverage the modality distribution captured by the model itself to construct the modality diffusion process. 

\begin{figure*}[!t]
\centering
\includegraphics[width=0.9\linewidth]{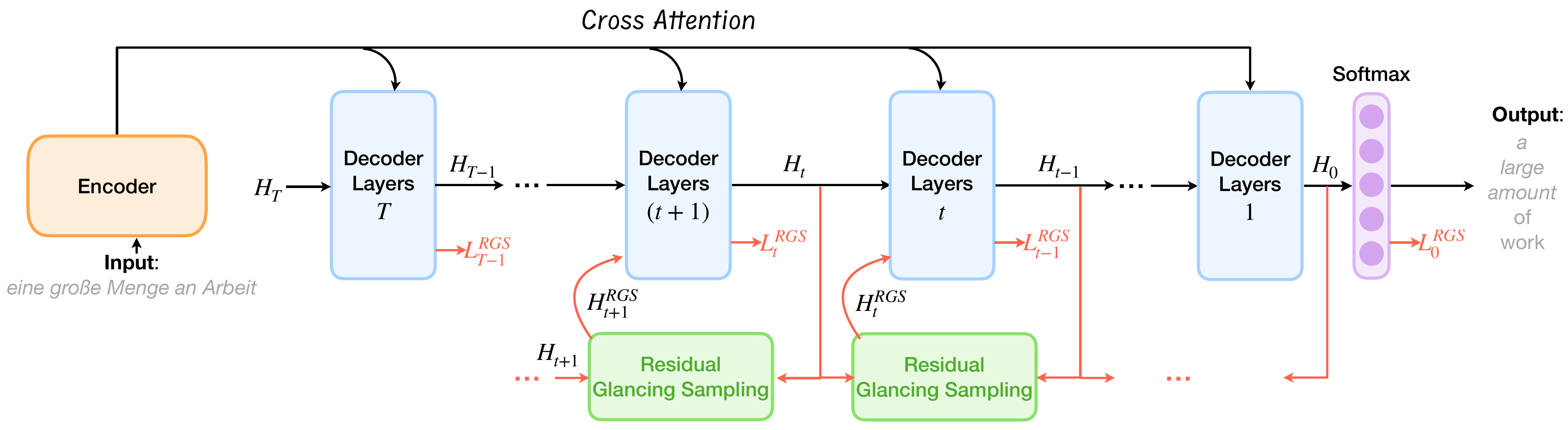}
\caption{Learning procedure of the layer-wise residual glancing sampling.}
\label{fig:RGS}
\vspace{-0.5em}
\end{figure*}

Specifically, $q^\text{MDP}$ should guarantee that the difficulty of the assigned modality learning task is appropriate for each transition. 
Thus, we define each transition of the modality diffusion process as the interpolation of distribution for target $\gP(\vy_0)$ and the prediction distribution of the next step $\gP_{t-1}$.
Mathematically, the posterior $q(\bm{y}_{1:T}|\bm{y}_0,\bm{x})$ can be decomposed as:
\begin{equation}
\begin{aligned}
    q^\text{MDP}&(\bm{y}_{1:T}|\bm{y}_0,\bm{x}):=\\
 &q(\bm{y}_T|\bm{y}_0,\bm{x})\prod_{t=2}^{T}q(\bm{y}_{t-1}|\bm{y}_t,\bm{y}_0,\bm{x})
\end{aligned}
\end{equation}
For simplicity, we omit $\bm{x}$ for $q$ in subsequent elaboration.
With the model $p_\theta$ and the target $\bm{y}_0$, we interpolate $\bm{y}_0$ and $p_\theta(\bm{y}_{T-1}|\bm{y}_{T},\bm{x})$ to construct intermediate distributions whose numbers of modalities are between the modalities in $\bm{y}_0$ and those captured by $p_\theta$:
\begin{equation}
    \label{eqn:q(YT|Y0)}
    \begin{aligned}
    %:= f(\gP_{(t-1)}, \vy_0; \gamma_{t})\\
    & q(\bm{y}_t|\bm{y}_0):=\bigg(\gamma_{t} \bm{1} + (1-\gamma_{t}) \bm{y}_0\bigg)\odot \gP_{t-1}/Z_{t} \\
    & :=\bigg(\gamma_{t} \gP_{t-1} + (1-\gamma_{t}) \gP_{t-1} \odot \bm{y}_0\bigg)/Z_{t},
    % & \text{where} \quad \gP_{t-1}=p_\theta(\bm{y}_{t-1}|\bm{y}_{t},\bm{x})
    \end{aligned}
\end{equation}
where $\gP_{t-1}=p_\theta(\bm{y}_{t-1}|\bm{y}_{t},\bm{x})$, $\bm{y}_0$ is the one-hot vector sequence of the target in data, $\gamma_t \in (0,1]$ is the hyper-parameter for controlling the interpolation, $\odot$ represents element-wise multiplication and $Z_{t}\in (0,1]^n$ is the factor for normalization.\footnote{Because $\sum_{\bm{y}_t}q(\bm{y}_t|\bm{y}_0)=1$ is not guaranteed after interpolation, we normalize the distribution with $Z_t=\sum_{w=1}^{|V|} \dot{q}(\bm{y}_t^w|\bm{y}_0)$, where $\dot{q}(\bm{y}_t|\bm{y}_0)=\gamma_{t} \gP_{t-1} + (1-\gamma_{t}) \gP_{t-1} \odot \bm{y}_0$, $|V|$ is the size of token categories and $\bm{y}_t^w$ is a sequence with the token $w$ on all the positions.}
We can consider the definition of Eq.~\ref{eqn:q(YT|Y0)} as data corruption on the modality probability landscape of $\gP$ by extracting the $\gP$ term out of the parenthesis.
% By extracting the $\gP$ term out of the parenthesis in Eq.~\ref{eqn:q(YT|Y0)}, the definition can be considered as data corruption on the modality probability landscape of $\gP$.
To ensure the form in Eq.~\ref{eqn:q(YT|Y0)} holds for $t\geq 1$~(See Appendix~\ref{sec:consistency}), we define $q(\bm{y}_{t-1}|\bm{y}_t,\bm{y}_0)$ as:
\begin{equation}
\label{eqn:transition}
    \begin{aligned}
    %& q(\bm{y}_{t-1}|\bm{y}_t,\bm{y}_0) = \gamma p_{t-2} + (1-\gamma) p_{\bm{y}_0}\odot p_{t-2},\\
    %q(\bm{y}_{t-1}|\bm{y}_t,\bm{y}_0) & = p_{t-2} + p_{\bm{y}_0}\odot (\sqrt[\gamma]{p_{t-2}}-p_{t-2}),\\
    % q(\bm{y}_{t-1}|\bm{y}_t,\bm{y}_0) & = \gamma_{t-1} p_{t-2} + (1-\gamma_{t-1}) p_{t-2} \odot \bm{y}_0
    %  + \frac{(1-\omega_{t-1})\gamma_{t-1}p_{t-2}}{(1+(1-\omega_{t})\gamma_{t})p_{t-1}}\bm{y}_0 \odot \bm{y}_t
    %:= f(\gP_{(t-1)}, f(\gP_{(t-2)}, \vy_0;\gamma_{t-1}), \omega_{t})\\ &
    q&(\bm{y}_{t-1}|\bm{y}_t,\bm{y}_0) :=\gamma_{t-1} \gP_{t-2} + \bigg((1-\gamma_{t-1})\\
    &\gP_{t-2}-\omega_{t}\gP_{t-1}\bigg) \odot \bm{y}_0
     + \omega_{t}Z_t \bm{y}_0 \odot \bm{y}_t
    \end{aligned}
\end{equation}
Here, $\omega_t$ is a hyper-parameter. Similarly, we also re-normalize $q(\bm{y}_{t-1}|\bm{y}_t, \bm{y}_0)$. Intuitively, the sum of the first term with $\gP_{t-2}$ and the second term with $\bm{y}_0$ performs the modality interpolation as in Eq.~\ref{eqn:q(YT|Y0)}, and adding the third term with $\bm{y}_0 \odot \bm{y}_t$ attempts to preserve the modalities captured by $\bm{y}_t$ in the denoising transitions.
%and $p_{t-2}=p_\theta(\bm{y}_{t-2}|\bm{y}_{t-1},\bm{x})$, $q(\bm{y}_{t-1}|\bm{y}_t,\bm{y}_0)$ incorporates the modalities included in $\bm{y}_t$ and part of the modalities in $\bm{y}_0$ that are likely to be captured by $p_{t-2}$. 
Thus, $\bm{y}_{t-1}$ serves as a intermediate target, with fewer modalities than $\bm{y}_0$ but more modalities than $\bm{y}_t$, for smoothing the modality learning.

% To train the non-autoregressive model with our modality diffusion process, we define the generative process $p_\theta(\bm{y}_{0:T}|\bm{x}):=p(\bm{y}_T|\bm{x})\prod_{t=1}^T p_\theta(\bm{y}_t|\bm{y}_{t-1},\bm{x})$ with the neural network. The non-autoregressive generative process uses the intermediate hidden representation to compute the softmax distribution of $\bm{y}_t$. For training, $p_\theta^\text{RGS}$ employs an improved glancing sampling strategy to reduce the difficulty of learning modality. In inference, we keep the decoding process of non-autoregressive models unchanged and decode the discrete tokens only once.

% Given the well scheduled modalities among NAR layers, it is also crucial to \zhouh{add some analysis why we need to propose RGS?}

\subsection{Non-autoregressive Generative Process}
We define the generative process similar to the iterative refinement process in NAR. Specifically, we use a generative process that only conditions on the input computed by the model itself for every step, removing the gap between training and inference. 
Besides, we also augment the training with residual glancing sampling~(RGS) to further smooth the modality learning, which improves over the glancing training~\citep{glat2021qian} by layer-wise sampling and selecting tokens that are not captured in the input.

\paragraph{Self-Conditioned Generative Process}
To keep consistency between the training and inference, we parameterized $p_\theta(\bm{y}_{t-1}|\bm{y}_t)$ completely by the model itself.
We set the decoder input sequence as the initial state $\bm{y}_T$, and the embedding of decoder inputs $emb(\bm{y}_T)$ to be the initial representation $\bm{H}_T$. For the generative transitions $p_\theta(\bm{y}_{t-1}|\bm{y}_t)$, we compute the next representation $\bm{H}_{t-1}$ with $\bm{H}_t$, and maps the hidden states representation to the softmax distribution for $p_\theta(\bm{y}_{t-1}|\bm{y}_{t},\bm{x})$:
\begin{equation}
\begin{aligned}
 & p_\theta(\bm{y}_{t-1}|\bm{y}_{t},\bm{x}) = \text{softmax} (\bm{H}_{t-1} \bm{V}^T), \\
 & \text{where} \quad \bm{H}_{t-1}=f_{\theta}^{(t)}(\bm{H}_t,\text{Enc}(\bm{x})),
\end{aligned}
\end{equation}
Here, $\text{Enc}$ is the encoder that maps the input $\bm{x}$ to hidden states, $\bm{V}$ is the vocabulary embedding matrix, and $f_\theta^{(t)}$ is parameterized by neural layers. Since $\bm{H}_t$ contains the information for $\bm{y}_t$ and is consistent in both training and inference, we use $\bm{H}_t$ as the input for $p_\theta(\bm{y}_{t-1}|\bm{y}_{t},\bm{x})$.
The self-conditioned generative process shares similar motivation with self-conditioning~\citep{chen2022analog} and step-unroll~\citep{savinov2022stepunrolled}, but operates in a purer way as we directly transits the hidden states as in typical decoders. 
% Depending on whether the parameters are shared across time $t$, we can create fully non-autoregressive models or iterative models. For fully non-autoregressive models that only forward the neural network once, $\theta_t$'s are not shared and the neural layers are stacked together. For iterative models, the parameters of $\theta_t$ are shared across timestep $t$.

\begin{figure}[t]
\begin{center}
\vspace{-1em}
\includegraphics[width=0.9\linewidth]{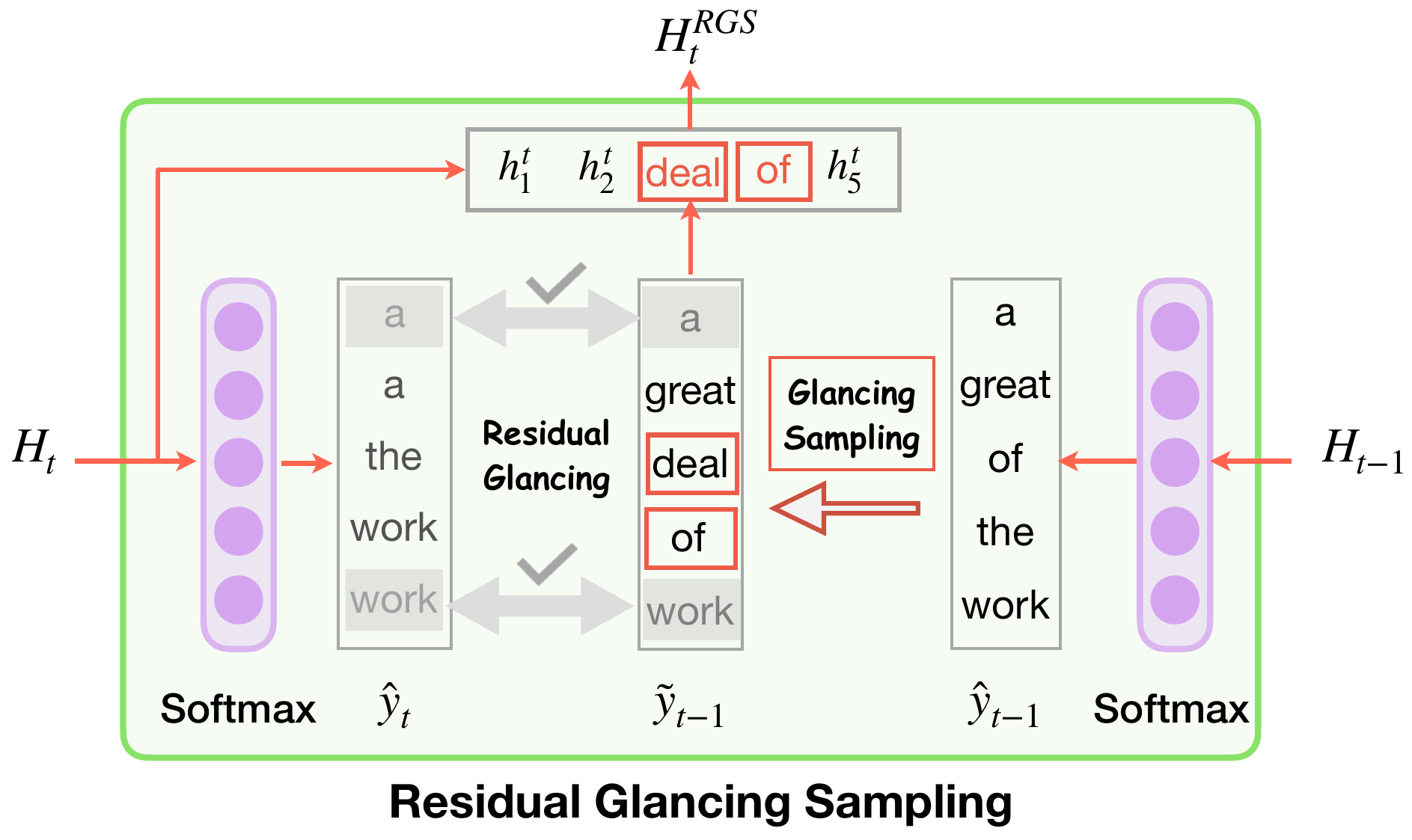}
\caption{The details of residual glancing sampling.}
\label{fig:RGS_details}
\vspace{-2em}
\end{center}
\end{figure}
\paragraph{Residual Glancing Sampling}
Previous work demonstrate that adaptive sampling target tokens according to the model performance can improve the generation quality of NAR~\citep{glat2021qian}.
For models training with diffusion processes, we strengthen the glancing training in a fine-grained way by layer-wise glancing and glancing remaining tokens. The overall residual glancing training procedure is shown in Figure~\ref{fig:RGS}.

For layer-wise glancing, we replace part of intermediate hidden states $\bm{H}_{t}$ with the embedding of tokens sampled from the intermediate target $\tilde{\bm{y}}_{t-1}=\arg \max q(\bm{y}_{t-1}|\bm{y}_t,\bm{y}_0)$, and predict the remaining target tokens $\overline{\tilde{\bm{y}}_{t-1}^\text{RGS}}$ using the glancing target tokens $\tilde{\bm{y}}_{t-1}^\text{RGS}$:
\begin{equation}
\begin{aligned}
    \mathcal{L}_{t-1}^\text{RGS} & = -\log p_\theta^\text{RGS}(\bm{y}_{t-1}|\bm{y}_t,\bm{x}) \\ 
    & = -\log p_\theta(\overline{\tilde{\bm{y}}_{t-1}^\text{RGS}}|\tilde{\bm{y}}_{t-1}^\text{RGS},\bm{y}_t,\bm{x}) 
\end{aligned}
\end{equation}
The sampling number of glancing tokens $\tilde{\bm{y}}_{t-1}^{RGS}$ is determined by the distance between the model output $\hat{\bm{y}}_{t-1}= \arg\max p_\theta(\bm{y}_{t-1}|\bm{y}_{t},\bm{x})$ and the intermediate target $\tilde{\bm{y}}_{t-1}$:
    $S_{t-1} = \alpha \cdot d(\tilde{\bm{y}}_{t-1},\hat{\bm{y}}_{t-1})$,
where $\alpha$ is a hyper-parameter for adjusting the sampling number.
Details for glancing could be found in \citet{glat2021qian}.
%We compare the output $\hat{\bm{y}}_{t-1}=\arg\max p_\theta(\bm{y}_{t-1}|\bm{y}_t,\bm{x})$ and the intermediate target $\bm{y}^*_{t-1}$ for computing the number of sampling tokens. 

% \zhouh{please combine the following paragraph with the above one.}
% For the generative transitions, we propose residual glancing to modify the selection strategy for glancing tokens to facilitate effective token sampling.
Besides layer-wise glancing, we propose to modify the scope for selecting glancing tokens as in Figure~\ref{fig:RGS_details}. Since the input captures part of the modalities, the model can learn only the remaining part that are not correctly captured in the input.
Thus, we utilize $\hat{\bm{y}}_{t}$ to schedule the learning, and modify the glancing sampling to only sampling $S_{t-1}$ tokens that are different between $\hat{\bm{y}}_t$ and $\tilde{\bm{y}}_{t-1}$:
\begin{equation}
\label{eqn:sampling}
    \mathbb{RGS}_{t-1}(\tilde{\bm{y}}_{t-1}) = Random(\tilde{\bm{y}}_{t-1}/\hat{\bm{y}}_{t})
\end{equation}
In residual glancing training, we randomly sample $\tilde{\bm{y}}_{t-1}^{RGS}$ from $\tilde{\bm{y}}_{t-1}$ with the operation $\mathbb{RGS}_{t-1}$.
%As we sample the tokens that have not been captured by previous transitions, the glancing training also divides the modalities to be learned in different transitions, which we find empirically effective in extensive experiments. 
% Besides, to narrow the gap between $\bm{H}_{t-1}$ and $\bm{y}_{t-1}$ for subsequent computation of $p_\theta^{RGS}$, we propose glancing schedule to fuse part of $\bm{y}^{*RGS}_{t-1}$ into $\bm{H}_{t}$ to compute $\bm{H}_{t-1}$, and reduce the fusion amount of $\bm{y}^{*RGS}_{t-1}$ gradually. Different from computing the loss $L_{t-1}^\text{RGS}$ with the residual glancing sampling, the glancing schedule aims to close the gap between $\bm{H}_{t-1}$ and $\bm{y}_{t-1}$ for the next transition $p_\theta(\bm{y}_{t-2}|\bm{y}_{t-1},\bm{y}_0)$. Specifically, the glancing schedule uses fewer glancing tokens than that of $\bm{H}_t^{RGS}$ and the number of tokens glancing schedule is computed by: $\mu S_{t-1}$, where $\mu$ is a hyper-parameter.

% \paragraph{Glancing Schedule}
% \qlh{Todo: The process for reducing the gap between $\bm{y}_t$ and output of $p_\theta(\bm{y}_t|\bm{y}_{t+1})$}

%And for DA-Transformer~\citep{dag2022huang}, we select the output tokens from the argmax decoding tokens with the link between output nodes. 
%In our implementation, we feed the output hidden states rather the intermediate decoding result to the next layer, which allow the training to be performed in an end-to-end fashion.

\subsection{Implementation Details}
\label{sec:training}
In this section, we will introduce the implementaion details for \method. To enhance the ability of modality capturing, we employ the DA-Transformer~\citep{dag2022huang}. The DA-Transformer~(DAT) can strengthen each transition of \method by distributing the modalities to different positions of expanded output sequences. We also elaborate the difference between the decoding iterations and the diffusion transitions in the parameterization part.

% With our modality diffusion process $q^\text{MDP}$ and non-autoregressive process $p_\theta^\text{RGS}$, we can train the model by optimizing the variational lower bound according to Eq.~\ref{eqn:vlb}:
% \begin{equation}
% \begin{aligned}
% &L = \sum_{t=1}^{T} L_{t-1}^\text{RGS} \quad \text{where} \quad L_{t-1}^\text{RGS}=D_{\text{KL}}(\\
% &q^\text{MDP}(\bm{y}_{t-1}|\bm{y}_t, \bm{y}_0)||p_\theta^\text{RGS}(\bm{y}_{t-1}|\bm{y}_t,\bm{x})). 
% \end{aligned}
% \end{equation}
% Since our diffusion process adds the modalities in the data gradually, each transition in the generative process only learns part of the modalities, which greatly reduce the number of modalities to learn for each transition. The residual glancing sampling strategy is only used in training without modifying the inference.
% For inference, our method keep the non-autoregressive generation unchanged, because the next prediction takes the last hidden representation rather than the intermediate discrete output as input.

\paragraph{DA-Transformer}
% Following previous work for NAR, we adopt the Transformer architecture for conditional sequence generation.
% To strengthen the performance, we can combine \method with other methods, such as CTC~\citep{ctc2006graves} and DAT~\citep{dag2022huang}.
%For CTC, we set the decoding length to be twice the source length. For DAT, the decoding length is set to be 8 times the source length for non-iterative models and 4 times for iterative models.
In training, the DAT model expands the output lengths to the predefined max length, and maps target tokens to multiple positions in the expanded sequence.
Since the output length of DAT is not equal to the length of the target $\bm{y}$, we use the best alignment of $\tilde{\bm{y}}_t$ as the glancing target and for computing $q^\text{MDP}$. The best alignment is obtained by maximizing the output probability:
\begin{equation}
\label{eqn:best_align}
    \bm{y}_{t}^{align} = \arg\max_{\bm{a} \in \Gamma(\tilde{\bm{y}}_t)} p_{t}(\bm{a}|\bm{x}),
\end{equation}
where $\Gamma$ expands $\bm{y}$ to the expanded output length by inserting blanks.
For the intermediate target $\tilde{\bm{y}}$ to compute loss $L_{t-1}^\text{RGS}$, we use the decoding result with the original target length $|\bm{y}|$ rather than the aligned target. 
In inference, after parallel neural network computation, DAT uses links between positions to extract output tokens in the expanded sequence, where we use the Joint-Viterbi decoding proposed by ~\citet{shao2022viterbi}.

 \begin{table*}[!t]
\begin{center}
\begin{small}
%\begin{sc}
\resizebox{0.85\linewidth}{!}{
\setlength{\tabcolsep}{1mm}{
\begin{tabular}{ll|c|>{\hspace*{2.5mm}}r@{}r@{}l@{}l|>{\hspace*{2.5mm}}r@{}r@{}l@{}l|>{\hspace*{2.5mm}}r@{}r@{}l@{}l|>{\hspace*{2.5mm}}r@{}r@{}l@{}l|>{\hspace*{1mm}}r@{}r@{}l@{}l|>{\hspace*{2mm}}r@{}l}
\toprule
\multicolumn{2}{l|}{\multirow{2}{*}{\bf Model}} & \bf Iter & \multicolumn{4}{c|}{\bf WMT14} & \multicolumn{4}{c|}{\bf  WMT14} & \multicolumn{4}{c|}{\bf WMT17} & \multicolumn{4}{c|}{\bf WMT17} & \multicolumn{4}{c|}{\bf Average} & \multicolumn{2}{c}{\multirow{2}{*}{\bf  Speedup}}\\
& &  & \multicolumn{4}{c|}{\bf En-De}  & \multicolumn{4}{c|}{\bf De-En} & \multicolumn{4}{c|}{\bf En-Zh} & \multicolumn{4}{c|}{\bf Zh-En}  &  \multicolumn{4}{c|}{\bf Gap} & \\
\midrule
\multirow{1}{*}{AR} 
%& Transformer~{\scriptsize \cite{transformer2017vaswani}} & $M$ & & 27&.6 &  & & 31&.4 & & &  34&.3 & & & 23&.7 & & & \tempneg0&.35 & & 1&.0x \\
& Transformer \textit{base} (Ours) & $M$ & & 27&.18* & & & 31&.48* & & & 34&.65* &  & & 23&.39*  & & & 0& &  & 1&.0x \\
%& Transformer \textit{big} (Ours) & $M$ & & 28&.69* & & & 32&.64* & & & -&- &  & & -&-  & & & -&- &  & 0&.9x \\
\midrule
\multirow{4}{*}{Iterative Models} 
%& Diff-LM~{\scriptsize \cite{li2022diffusion}} & 20 & & 15&.33 &  & & 17&.31 &  & & \multicolumn{2}{c}{-} &  & & \multicolumn{2}{c}{-}  &  & & \tempneg11&.60$^\diamondsuit$ &  & 1&.3x\\
& Diff-LM~{\scriptsize \cite{li2022diffusion}} & 20 & & 17&.41 &  & & 19&.69 &  & & \multicolumn{2}{c}{-} &  & & \multicolumn{2}{c}{-}  &  & & \tempneg10&.78$^\diamondsuit$ &  & 0&.6x\\
& CDCD~{\scriptsize \cite{dieleman2022continuous}} & 100 &  & 20&.0 & & & 26&.0 &  & & \multicolumn{2}{c}{-} &  & & \multicolumn{2}{c}{-} &  & &  \tempneg6&.33$^\diamondsuit$ & & -&-\\
& Difformer~{\scriptsize \cite{gao2022difformer}} & 20 & & 23&.80 & & & \multicolumn{2}{c}{-} &  & & \multicolumn{2}{c}{-} &  & & \multicolumn{2}{c}{-} &  & & \tempneg3&.38$^\diamondsuit$ &  & -&-\\
& \textsc{DiNoiSer}~{\scriptsize \cite{ye2023dinoiser}} & 20 & & 24&.26 & & & 29&.05 &  & & \multicolumn{2}{c}{-} &  & & \multicolumn{2}{c}{-} &  & & \tempneg2&.68$^\diamondsuit$ &  & -&-\\
% \midrule
% \multirow{1}{*}{Iterative NAR} 
% & CMLM~{\scriptsize \cite{cmlm2019ghazvininejad}} & 10 & & 24&.61 &  & & 29&.40 &  & & \multicolumn{2}{c}{-} &  & & \multicolumn{2}{c}{-}  &  & & \tempneg2&.88$^\diamondsuit$ &  & 2&.2x\\
%& SMART~{\scriptsize \cite{smart2020}} & 10 & & 25&.10 & & & 29&.58 &  & & \multicolumn{2}{c}{-} &  & & \multicolumn{2}{c}{-} &  & & \tempneg2&.55$^\diamondsuit$ &  & 2&.2x\\
% DisCo~{\scriptsize\cite{disco2020kasai}} & $\approx$4 & & 25&.64 &  & & \multicolumn{2}{c}{-} &  & & \multicolumn{2}{c}{-} &  & & \multicolumn{2}{c}{-} &  & & \tempneg2&.31$^\diamondsuit$ &  & 3&.5x \\
% & Imputer~{\scriptsize\cite{imputermt2020saharia}} & 8 & & 25&.0 & & & \multicolumn{2}{c}{-} &  & & \multicolumn{2}{c}{-} &  & & \multicolumn{2}{c}{-} & & & \tempneg2&.96$^\diamondsuit$ &  & 2&.7x \\
& SUNDAE~{\scriptsize\cite{savinov2022stepunrolled}} & 10 & & 25&.99 &  & & 30&.24 &  & & \multicolumn{2}{c}{-} &  & & \multicolumn{2}{c}{-} &  & & \tempneg1&.36$^\diamondsuit$ &  & 2&.2x \\
& DAT+Iterative refinement$^\dag$ & 3 & & 25&.67* &  & & 30&.85* &  & & \multicolumn{2}{c}{-} &  & & \multicolumn{2}{c}{-} &  & & \tempneg1&.07$^\diamondsuit$ &  & 7&.2x \\
% & CMLMC~{\scriptsize\cite{cmlmc2021}} & 10 & & 26&.40 &  & & 30&.92 &  & & \multicolumn{2}{c}{-} &  & & \multicolumn{2}{c}{-} &  & & \tempneg1&.23$^\diamondsuit$ &  & 1&.7x \\
\midrule
\multirow{4}{*}{Non-iterative Models} 
%& Vanilla NAR~{\scriptsize\cite{nat2018gu}} & 1 & & 11&.79 &  & & 16&.27 & & & 18&.92 & & & 8&.69 &  & & \tempneg15&.68 &  & 15&.3x \\
& CTC~{\scriptsize\cite{ctc2018libovicky}} & 1 & & 17&.73* &  & & 21&.48* & & & 25&.77* & & & 12&.33* &  & & \tempneg9&.85 &  & 14&.3x \\
% A\bm{x}E~{\scriptsize\cite{axe2020ghazvininejad}} & 1 & & 20&.40 &  & & 24&.90 & & & \multicolumn{2}{c}{-} &  & & \multicolumn{2}{c}{-} & & & \tempneg7&.24$^\diamondsuit$ &  & 14&.2x \\
% & GLAT~{\scriptsize\cite{glat2021qian}} & 1 & & 19&.42 &  & & 26&.51 &  & & 29&.79 & & & 18&.88 &  & & \tempneg5&.95 &  & 15&.3x \\
% & OaXE~{\scriptsize\cite{oaxe2021du}} & 1 & & 22&.4 & & & 26&.8 &  & & \multicolumn{2}{c}{-} &  & & \multicolumn{2}{c}{-} &  & & \tempneg5&.28$^\diamondsuit$ &  & 14&.2x \\
& GLAT+CTC~{\scriptsize\cite{glat2021qian}} & 1 & & 24&.85* &  & & 28&.37* &  & & 30&.20* &  & & 17&.57* &  & & \tempneg3&.92  &  & 14&.3x \\
% CTC + DSLP~{\scriptsize\cite{dslp2021huang}} & 1 & & 24&.81 &  & & 28&.33 &  & & \multicolumn{2}{c}{-} &  & & \multicolumn{2}{c}{-} &  & &  \tempneg3&.32$^\diamondsuit$ & & 14&.0x \\
& DAT $^\dag$~{\scriptsize\cite{dag2022huang}} & 1 & & 26&.47* &  & & 30&.22* &  & & 33&.27* &  & & 23&.21* &  & & \tempneg0&.88 &  & 13&.0x \\
& DAT+DSLP $^\dag$ & 1 & & 26&.08* &  & & 30&.34* &  & & \multicolumn{2}{c}{-} &  & & \multicolumn{2}{c}{-} &  & & \tempneg1&.12$^\diamondsuit$ &  & 12&.6x \\
%+ BeamSearch & 1 & & 27&.02 & & & 31&.24 & & & 34&.21 &  & & 24&.22 & & & \tempneg0&.43 & & 7&.1x\\
\midrule
\multirow{2}{*}{Ours} 
%& \method~(CTC) \ \  & 1 & & 26&.46 & & & 30&.48 & & & 31&.77 &  & & 20&.87 &  & & \tempneg2&.21 &  & 14&.3x \\
%+ beamsearch\ \  & 1 & -&- & -&- & -&- & -&- & -&- & -&- & -&- & -&- & \tempneg-&- & \tempneg-&- & 14&.6x \\
& \method$^\dag$ \ \  & 1 & & 26&.72 & & & 31&.35 & & & 34&.12 & & & 23&.74 & & & \tempneg0&.19 & & 13&.0x \\
%+ beamsearch\ \  & 1 & & 27&.58 & & & 31&.95 & & & 34&.83 & & & 24&.81 & & & +0&.19 & & 7&.1x \\
& \method$^\dag$ \ \  & 3 & & 27&.91 & & & 31&.55 & & & 35&.09 & & & 24&.02 & & & +0&.47 & & 7&.2x \\
%& \method \textit{big}~(DAT)$^\dag$ \ \  & 3 & & 29&.28 & & & 32&.86 & & & -&- & & & -&- & & & -&- & & 6&.5x \\
\bottomrule
\end{tabular}
}}
%\end{sc}
\end{small}
\end{center}
\caption{Results on WMT14 En$\leftrightarrow$De and WMT17 Zh$\leftrightarrow$En. The average gap is computed against our Transformer implementation. * represents the results are obtained from our re-implementation, and $\diamondsuit$ indicates that the average gap is only computed with available results. For models with $\dag$, we use the Joint-Viterbi decoding proposed by~\citet{shao2022viterbi} for inference. The average gap is computed against the results of our implemented Transformer \textit{base}.}
\label{tab:main_result}
\vspace{-1em}
\end{table*}

\paragraph{Parameterization}
As each transition of the generative process directly forward the hidden states to the next transition, we can build fully non-autoregressive models by stacking the neural layers or iterative models by sharing the parameters of $\theta_t$.
%For fully non-autoregressive models, our method keeps the inference unchanged.
And we can use part of the layers in the decoder to perform one denoising transition so that one forward inference of the decoder can perform multiple denoising transitions. 
%If one transition is parameterized with 3 decoder layers, then a decoder of 6 layers can perform 2 generative transitions with 1 iteration.

\paragraph{Training}
% To reduce the influence of multi-modality in each term $L_{t-1}^\text{RGS}$, we optimize the model using a single target instead of the target distribution $q(\bm{y}_{t-1}|\bm{y}_t, \bm{y}_0)$. 
Since $q(\bm{y}_{t-1}|\bm{y}_t, \bm{y}_0)$ is a distribution consists of multiple modalities, directly minimizing the KL-divergence in Eq.~\ref{eqn:our_vlb} introduces multi-modal targets for every source input. Thus, we optimize the model using the sequence with the highest probability: $\tilde{\bm{y}}_{t-1}=\arg \max q(\bm{y}_{t-1}|\bm{y}_t,\bm{y}_0)$. The KL-divergence can then be rewritten as:
\begin{equation}
   \mathcal{L}_{t-1}^\text{RGS}= -\log p_\theta^\text{RGS}(\tilde{\bm{y}}_{t-1}|\bm{y}_t,\bm{x})
\end{equation}
 % Since $q(\bm{y}_{t-1}|\bm{y}_t,\bm{y}_0)$ defined in Eq.~\ref{eqn:transition} is formed by $\bm{y}_0$ and $p_\theta(\bm{y}_{t-2}|\bm{y}_{t-1},\bm{x})$, we can leverage the next transition to find the next part of modalities suitable for $p_\theta(\bm{y}_{t-1}|\bm{y}_{t},\bm{y}_0)$ to fit. The process of gradually adding modalities guides the model to keep learning more modalities that are not captured. 

To reduce the denoising steps needed to achieve high quality, we use a small number for the total diffusion steps $T$ and attempt to fit more modalities with each generative transition $p_\theta(\bm{y}_{t-1}|\bm{y}_t,\bm{x})$.
For training efficiency, we sample a diffusion timestep $t$ to compute the loss for each training step.

\section{Experiments}
\label{sec:exp}
% \begin{itemize}
% \item settings
% \item Table 1 comparison with AT and NAR models
% \item Table 2 results with deeper encoder and decoder layers
% \item Table 3 results for different language
% \item Table 4 Ablation for hierarchical and diffusion
% \item Table 5 results with more times of decomposition
% \item Table 6 experiments for modality capturing in different layers
% \end{itemize}

Different from most of the previous work for NAR, we directly train our models on raw data without using data distilled from AR. To verify the effectiveness of our method, we compare \method with several strong NAR and AR baselines on several sequence generation tasks. 
%And we further conduct experiments in the setting of deep models with more neural layers to exploit the potential of \method. 
Furthermore, analysis and ablation studies study are also conducted to demonstrate the effects of each component.
\subsection{Experimental Settings}
\paragraph{Benchmarks}
We conduct experiments on machine translation, paraphrase generation and image captioning. For machine translation, we use 10 machine translation benchmarks: WMT14 En$\leftrightarrow$De, WMT17 En$\leftrightarrow$Zh, WMT14 En$\leftrightarrow$Fr, WMT16 En$\leftrightarrow$Ro and WMT13 En$\leftrightarrow$Es. The preprocessing follows the procedure in~\citet{kdnat2020zhou} and~\citet{disco2020kasai}. For paraphrase generation and image captioning, we use the Quora Question Pairs dataset\footnote{\url{https://www.kaggle.com/c/quora-question-pairs}}~(QQP), and MS-COCO dataset~\citep{lin2014microsoft} respectively. The data are tokenized and segmented into subwords using Byte-Pair Encoding~\citep{bpe2016sennrich}.

\paragraph{Evaluation Metrics}
We report the sacreBLEU\footnote{Except 'tok.zh' for En-Zh, the signature is "BLEU+ case.mixed+numrefs.1+smooth.exp+tok.13a+version.1.5.1"}~\citep{sacrebleu2018} scores for machine translation, and the tokenized bleu results are shown in Appendix~\ref{sec:ad_mt_results}. For speedups compared with the Transformer \emph{base}, we follow previous work~\citep{nat2018gu} by evaluating on WMT14 En-De with batch size 1. Further comparison for decoding speedup are provided in Appendix~\ref{sec:infer_speedup}.

\begin{table*}[t]
\centering
\begin{small}
\resizebox{0.9\linewidth}{!}{
\begin{tabular}{l|cc|cc|cc|cc|cc}
\toprule
\multirow{2}{*}{\textbf{Models}} &  \multicolumn{2}{c|}{\textbf{WMT14}} & \multicolumn{2}{c|}{\textbf{WMT17}} & \multicolumn{2}{c|}{\textbf{WMT14}} &
 \multicolumn{2}{c|}{\textbf{WMT16}} &  \multicolumn{2}{c}{\textbf{WMT13}} \\
 & \textbf{En-De} & \textbf{De-En} & \textbf{En-Zh} & \textbf{Zh-En} & \textbf{En-Fr} & \textbf{Fr-En} & \textbf{En-Ro} & \textbf{Ro-En} & \textbf{En-Es} & \textbf{Es-En}\\
\midrule
Transformer \textit{base} &  27.18 &	 31.48 & 34.65 & 23.39 & 38.26 &  35.70	&  33.91 & 33.70 & 33.53 & 34.19\\
 \method \textit{base} &  \textbf{27.91} & 31.55 & \textbf{35.09} & \textbf{24.02} & \textbf{38.73} &  \textbf{36.61} & 33.92 & 33.64 & \textbf{33.95} & \textbf{34.74} \\
 % \midrule
 % \textbf{Models} &    \textbf{WMT14 Fr-En} &  \textbf{WMT16 En-Ro} &  \textbf{WMT16 Ro-En} &  \textbf{WMT13 En-Es} &  \textbf{WMT13 Es-En}\\
 \midrule
 Transformer \textit{big} & 28.01 & 32.11 & 36.31 & 23.60 & 40.16 & 37.90 & 33.46& 32.68& 34.64& 35.01\\
 \method \textit{big} & \textbf{28.62} & \textbf{32.32} & 36.21 & \textbf{24.40}& 40.12 & 37.87& \textbf{34.34} & \textbf{33.65} & 34.71 & \textbf{35.40}\\
 % Transformer &  36.13	&  34.44 & 34.20 & 33.63 & 34.53\\
 % \method &  36.99 & 34.37 & 34.24 & 34.08 & 35.11 \\
\bottomrule
\end{tabular}}
\end{small}
\caption{BLEU scores on the 10 Machine Translation Benchmarks}
\label{tab:translation}

\vspace{-1em}
\end{table*}

\begin{table*}[t]
\begin{minipage}{0.35\linewidth}
\centering
\includegraphics[width=0.84\linewidth]{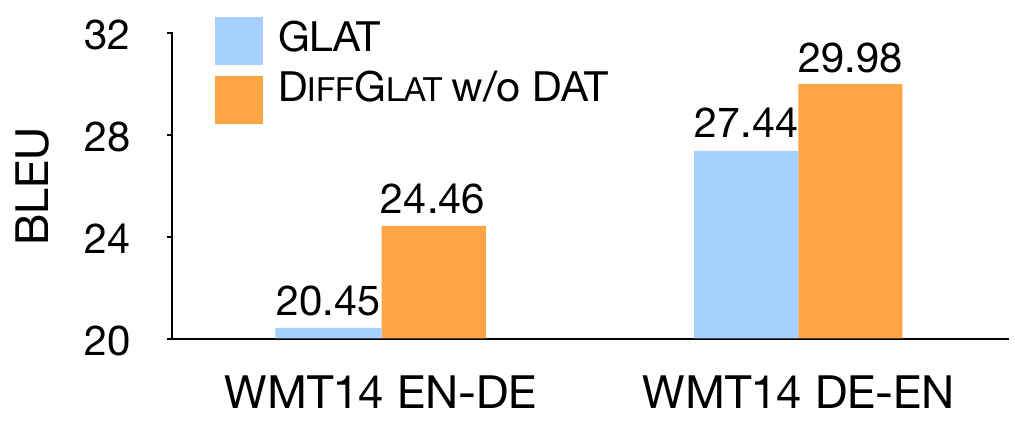}
\captionof{figure}{Results with GLAT}
\label{tab:more_base}
\end{minipage}
\begin{minipage}{0.65\linewidth}
\centering
\resizebox{\linewidth}{!}{
\begin{small}
\begin{tabular}{l|cc|cccc}
\toprule
& \multicolumn{2}{c|}{\textbf{QQP}} & \multicolumn{4}{c}{\textbf{MS-COCO 2014}}\\
& BLEU4 & ROUGE-L & BLEU4 & METEOR & ROUGE-L & CIDEr\\
\midrule
Transformer & 28.13  &58.19 & 34.0 & 28.1 & 56.0 & 112.3\\
DAT & 28.82 & 59.76 & 33.5& 27.0 & 56.5 & 106.5\\
\method & 29.86 & 60.23 & 34.9 & 28.0 & 56.7 & 112.5\\
\bottomrule
\end{tabular}
\end{small}}
\caption{Performance on paraphrasing and image captioning.}
\label{tab:gen}
\end{minipage}
% \hfill
% \begin{minipage}{0.36\linewidth}
% \centering
% \includegraphics[width=0.84\linewidth]{figure/ablation_RGS2.pdf}
% % \centering
% % \resizebox{0.85\linewidth}{!}{
% % \begin{small}
% % \begin{tabular}{l|cc}
% % \toprule
% % \multirow{2}{*}{}& WMT14 & WMT14 \\
% % & En-De & De-En \\
% % \midrule
% % \method & 28.57 & 32.08\\
% % w/o residual glancing & 28.01 & 31.65\\
% % w/o glancing schedule & 28.14 & 31.77\\
% % % w/o DAT & 25.17 & 29.26\\
% % \bottomrule
% % \end{tabular}
% % \end{small}
% % }
% \captionof{figure}{Residual glancing ablation}
% \label{fig:glancing}
% \end{minipage}
\vspace{-1em}
\end{table*}

\paragraph{Hyper-parameters}
We build our models based on the Transformer~\citep{transformer2017vaswani} architecture and use the Transformer \emph{base} setting as default. 
%We also conduct experiments with the Transformer \emph{big} setting for more extensive experiments. For machine translation, 
%the learning rate warms up to 5e-4 in the first 10k steps and decays with inverse square root schedule. 
For machine translation, we train all the models, including the autoregressive Transformer, with batches of 64k tokens and 300k training steps using the Adam optimizer~\citep{adam}. The training of \method with iterative decoding takes about 73 hours on 16 NVIDIAA100-80G GPUs. We average the best 5 checkpoints for BLEU scores on the validation set to get the final model.
%For the subsequent training steps, we train the model with the modality diffusion process.
All the models are built with 6 Transformer decoder layers, and we use 3 decoder layers for each transition $p_{\theta}(\bm{y}_{t-1}|\bm{y}_t,\bm{x})$. Thus, each decoding iterations performs 2 denoising diffusion steps.
%Correspondingly, we set the length of diffusion process $T$ to 6 for the model with 3 decoding iterations and $T=2$ for the non-iterative models, making training and inference consistent. 
For paraphrasing and image captioning, we use smaller architecture and shorter training steps. The detailed hyper-parameters can be found in Appendix~\ref{sec:hyper}.

% \begin{minipage}{0.25\linewidth}
% \centering
% \begin{small}
% \resizebox{\linewidth}{!}{
% \begin{tabular}{l|cc}
% \toprule
% \multirow{2}{*}{}& WMT14 & WMT14 \\
% & En-De & De-En \\
% \midrule
% GLAT & 20.45 & 27.44\\
% \method w/o DAT & \textbf{24.46} & \textbf{28.95}\\
% \midrule
% GLAT+CTC  & 24.85 & 28.37\\
% \method (CTC) & \textbf{25.92} & \textbf{29.98}\\
% \bottomrule
% \end{tabular}  
% }
% \end{small}
% \caption{\method }
% \label{tab:more_base}
% \end{minipage}

\subsection{Main Results}
From the results in Table~\ref{tab:main_result}, we can find that \method achieves considerable improvement over strong baselines. With a process that gradually adds modalities in the denoising pass, \method reduces the number of modalities to learn for each transition, enhancing the ability for capturing modalities in data. Depending on whether the model parameters are reused iteratively, /method can be trained for non-iterative or iterative decoding. For the non-iterative setting, \method only modifies the training procedure, thus can keep the inference the same as the process of the base model. Compared with several strong baselines, we highlight the advantages of \method:
\begin{itemize}
    \item Our method can achieve better generation quality than strong NAR or even AR. 
    %Averaging the scores on 4 benchmarks, our method improves 1.2 BLEU based on GLAT+CTC, and 0.7 BLEU based on DAT. 
    With only one iteration of parallel decoding, our model achives higher BLEU scores than that of previous non-autoregressive models. And when applying iterative decoding, our model can even outperform the Transformer with a margin of 0.47 BLEU on average. For comparison, directly applying iterative refinement to DAT does not improve the performance.
    %~\footnote{As the performance of the original DAT drops drastically with iterative decoding, we also train the DAT to perform iterative refinement.}
    \item \method overcomes the slow sampling of diffusion models and achieves high decoding efficiency. \method completely keeps the fast speed of parallel decoding for the non-iterative setting,  
    %For models reported in Table~\ref{tab:main_result}, we train the non-iterative \method using a diffusion process with only 2 steps, thus each step learns a complex multi-modal distribution to finish the generation efficiently. 
    and can still achieve a $7.2\times$ speedup with 3 decoding iterations. 
    % \item \method can be combined with various other methods. Since our approach only introduce a training procedure without changing architecture or inference process, we can apply our method to different models.
    \item We can also use \method without DAT or combine \method with CTC, where our approach achieves more improvements over the baselines. The results in Table~\ref{tab:more_base} show that our method improves 1.5$\sim$4 BLEU over GLAT. The CTC results are shown in Appendix~\ref{sec:ad_mt_results}.
\end{itemize}

\paragraph{Results on 10 Machine Translation Benchmarks}
To evaluate the performance more comprehensively, we compare the autoregressive Transformer with \method on 10 Benchmarks, and list the results in Table~\ref{tab:translation}. We conduct experiments for both the \textit{base} and the \textit{big} setting.
Our results show that \method outperforms or achieves comparable results to the Transformer on the 10 benchmarks.
% For those datasets where \method and Transformer performs comparably: WMT14 De-En, WMT16 En-Ro and WMT16 Ro-En, previous work on NAR finds smaller gap between AR and NAR on these datasets~\cite{cmlm2019ghazvininejad, glat2021qian}, indicating less multi-modality in those datasets. We argue that the less multi-modality in the data results in less improvements compared with the Transformer. 

\paragraph{Paraphrasing and Image Captioning}
Besides machine translation, we also conduct experiments for two other text generation tasks: paraphrase generation and image captioning. 
%On these two tasks, the models with DAT use Lookahead decoding~\citep{dag2022huang} for better performance. 
For paraphrase generation, \method can outperform the Transformer with about 1.7 BLEU scores and 2 ROUGE scores. As for image captioning, \method greatly improves over the DAT baseline and achieves slightly better results on the 4 metrics compared with the Transformer. Results on paraphrasing and image captioning shows that \method generalizes well for text generation tasks.

\subsection{Ablation Study}

\begin{figure*}[t]
\begin{minipage}{0.35\linewidth}
% \begin{table}[t!]
% \sbox\tmpbox{
\centering
\begin{small}
\resizebox{\linewidth}{!}{
\begin{tabular}{l|cc}
\toprule
\multirow{2}{*}{}& WMT14 & WMT14 \\
& En-De & De-En \\
\midrule
DAT & 25.96 & 30.02\\
+D3PM absorbing & 26.84 & 30.70\\
+D3PM uniform  & 27.36 & 28.14\\
%~{\tiny\cite{austin2021discrete}}
%+Modality Diffusion w/o $\bm{y}_0\odot \bm{y}_t$ & 27.22 & 31.38\\
+Modality Diffusion & 27.91 & 31.55\\
\bottomrule
\end{tabular}  
}
\end{small}

\captionof{table}{Comparison of different diffusion processes. All the models use the residual glancing training.}
\label{tab:process}
% \end{table}
\end{minipage}\hfill
\begin{minipage}{0.3\linewidth}
% \centering
% \includegraphics[width=\linewidth]{figure/iterative4.pdf}
% \caption{The BLEU score curves with iterative decoding.}
% \label{fig:iteration}
\centering
\includegraphics[width=\linewidth]{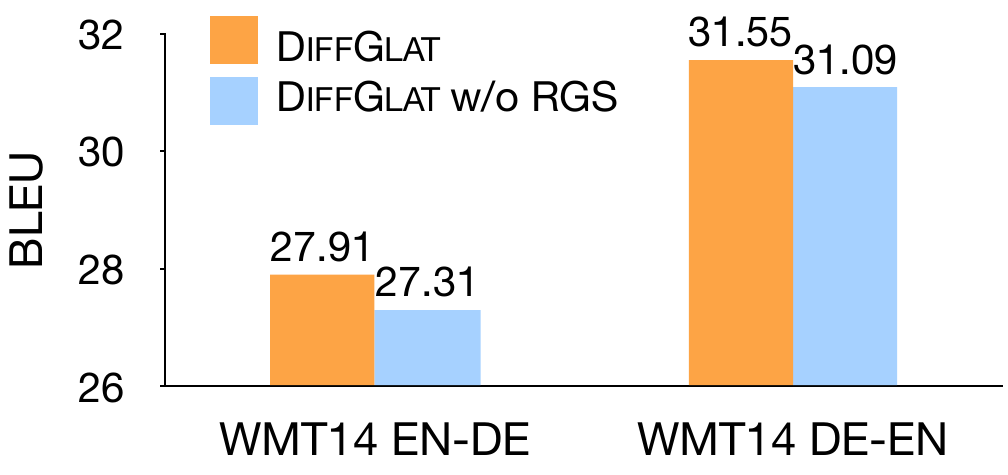}
\captionof{figure}{Ablation of the residual glancing strategy}
\label{fig:glancing}
\end{minipage}
\hfill
\begin{minipage}{0.3\linewidth}
\centering
\includegraphics[width=0.9\linewidth]{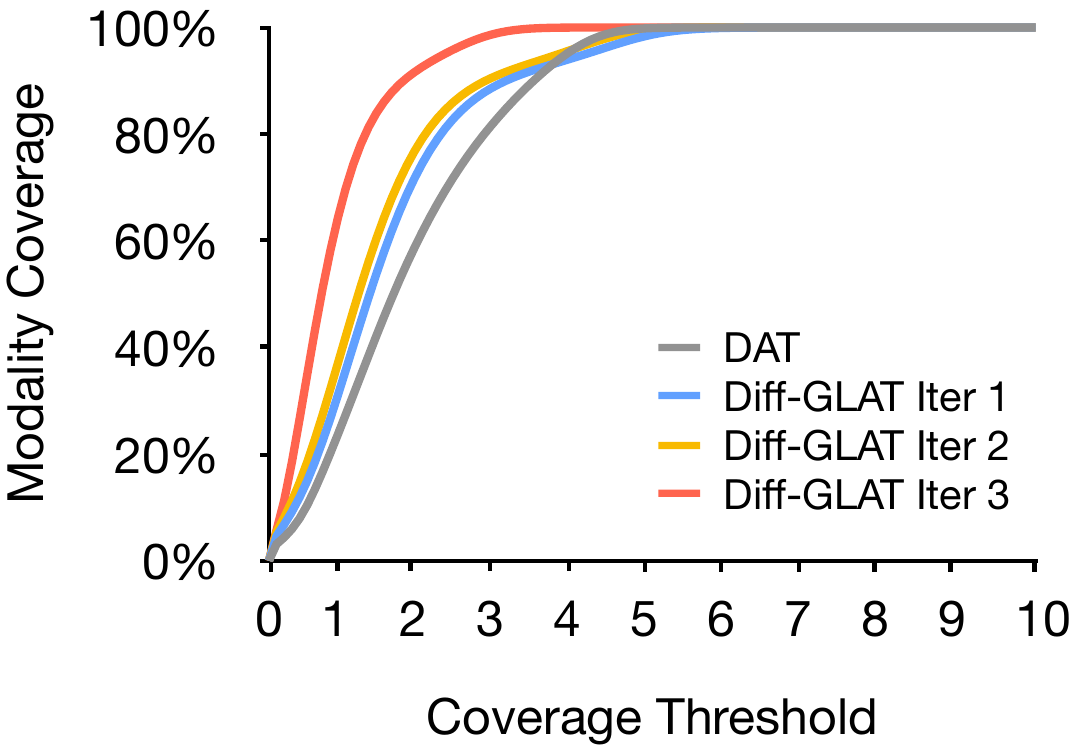}
\vspace{-0.5em}
\caption{The modality coverage curves on WMT14 En-De.}
\label{fig:coverage}
\end{minipage}
\vspace{-1em}
\end{figure*}

\paragraph{Comparison of Different Diffusion Processes}
We substitute the modality process in Section~\ref{sec:modality_diffusion} to compare the performance of different diffusion processes.
%as the process can be in various forms. 
For comparison, we use two typical discrete diffusion processes proposed in~\citep{austin2021discrete}: D3PM absorbing and D3PM uniform. In each forward step, the D3PM absorbing process masks each token of $\bm{y}_{t-1}$ with some given probabilities $\beta_t$, while the D3PM uniform process substitutes tokens in $\bm{y}_{t-1}$ with any other tokens uniformly. In our experiment, $\beta_t$ is set to $1/(T-t+1)$. From the results in Table~\ref{tab:process}, we can find that our modality gains significant improvement over the absorbing and uniform process.

\paragraph{Effectiveness of Residual Glancing Training}
To verify the effectiveness of the residual glancing, we remove our proposed modification for the glancing training for comparison. The results are shown in Figure~\ref{fig:glancing}, note that the result of \method without RGS uses the original glancing training. We can find that removing residual glancing causes a performance decline, indicating that our residual glancing improves the original glancing training.

\subsection{Analysis}

% \begin{figure}[t]
% \centering
% \includegraphics[width=0.7\linewidth]{figure/diff_iteration2.pdf}
% \caption{The effect of different diffusion process lengths.}
% \label{fig:diff_lengths}
% \end{figure}

% \begin{figure}[t]
% \centering
% \includegraphics[width=0.7\linewidth]{figure/modality_coverage2.pdf}
% \caption{The modality coverage on the WMT14 En-De training set under different coverage thresholds.}
% \label{fig:coverage}
% \end{figure}

% \paragraph{Performance with Iterative Decoding}
% The BLEU scores for different decoding iterations are presented in Figure~\ref{fig:iteration}. Here, the evaluated models are trained with generative processes of 3 iterations. We find that the BLEU scores increases as the number of iterations increases to 3, and the scores do not increase when the iterations exceeds 3.
% Note that the decoding iterations in inference can be different from that in training. Since the model uses a generation process of 3 iterations for training, the model also achieves the best scores with 3 decoding iterations, which is consistent with training.

\paragraph{Progressive Modality Capturing}
To measure the performance of modality capturing, we compute the modality coverage percentage under the corresponding thresholds, and the coverage curve is presented in Figure~\ref{fig:coverage}. Since the exact modalities for real world data is unavailable, we compute the coverage for the data point $(\bm{x},\bm{y})$ as the modality is covered when the related data points are covered. Specifically, we compute the normalized loss for each data point with the trained model: $L_\theta(\bm{x},\bm{y})=-\log p_\theta(\bm{y}|\bm{x})/|\bm{y}|$. And the data point is considered covered by the model if $L_\theta(\bm{x},\bm{y}) \leq \tau$, where $\tau \in \mathbb{R}$ is the threshold. For models with DAT, we compute $L_\theta(\bm{x},\bm{y})$ with the best aligned path $\bm{a}^* = \argmax_{\bm{a} \in \Gamma}P(\bm{y},\bm{a}|\bm{x})$. The curves depict the training data coverage under each threshold.

% By increasing the threshold gradually, we draw the curve for the coverage percentage of the training set in Figure~\ref{fig:coverage}.
In the figure, we can find that the curves of \method is overall on top of the DAT curve, indicating our model captures more modalities. And the three curves of \method shows that \method captures more modalities as the model performs more denoising steps.
% We measure the performance of modality capturing for different decoding steps by computing the KL-divergence between the data distribution and the output distribution of different iteration steps. Formally, the modality capturing accuracy~(MCA) for data $(\mathcal{\bm{x}},\mathcal{\bm{y}})$ is defined as: $\text{MCA}_{(\mathcal{\bm{x}},\mathcal{\bm{y}})}(p_\theta)=\mathbb{E}_{(\bm{x},\bm{y})\sim (\mathcal{\bm{x}},\mathcal{\bm{y}})}[-\log p_\theta(\bm{y}|\bm{x})]/|\bm{y}|$, where $|\bm{y}|$ is the length of target sequence.
% For models trained using objectives with alignment~(e.g. DAT and CTC), we compute MCA with the best alignment sequence computed by Eq.~\ref{eqn:best_align}.

% We compute MCA on the training set of WMT14 En$\leftrightarrow$De, and the results are shown in Figure~\ref{fig:MCA}. The lower MCA indicates the model learns the modalities in the data more accurately. As the decoding iterations increases, the MCA metric decreases and the BLEU score increases. Thus, \method captures the modalities gradually as the generative process proceeds. 
%Also, compared with the baseline DAT, \method achieves better performance for modality capturing.

\paragraph{Analysis for Decoding Iterations and Layers}
We also conduct experiments to analyze the performance with different decoding iterations and the effect of layer numbers for each denoising transition. The results are provided in Appendix~\ref{sec:appendix_analysis}.

\section{Related Work}
\label{sec:related_main}
The work for non-autoregressive neural machine translation can be divided into non-iterative and iterative methods.
%The non-iterative methods only forward the neural network once in generation, which keep high inference speed with parallel decoding. 
Previous non-iterative methods attempt to enhance the generation quality by earning from autoregressive models~\citep{imitatenat2019wei, guo2020fine}, incorporating light sequential decoding~\citep{crf2019sun, dag2022huang}, and introducing latent variables~\citep{pnat2019bao, flowseq2019ma, song2021alignart, bao2022latent}. Besides, models with iterative decoding have also been developed~\citep{iterativerefinement2018lee,cmlm2019ghazvininejad,levenshtein2019gu,dslp2021huang,imputermt2020saharia,savinov2022stepunrolled,cmlmc2021}.

\method also has connections with diffusion models for discrete data but differs from them in both probabilistic modelling and design motivation. More discussions can be found in Appendix~\ref{sec:related}.

\section{Conclusion}
\label{sec:conclusion}
In this work, we propose \method, a parallel sequence generation model trained with the modality diffusion process and residual glancing sampling. 
By smoothing the learning of modalities in the diffusion model framework, \method significantly improves the generation quality of parallel generation. 
Compared with the autoregressive Transformer, \method achieves superior performance in both accuracy and efficiency for multiple sequence generation tasks, demonstrating the potential of the parallel generation paradigm. 
% Note that the proposed \method has connections with continuous diffusion models and iterative non-autoregressive models for text generation, but \method differs from them in both probabilistic modeling and designing intuition. More details can be found in Appendix~\ref{sec:related}.

% \paragraph{Limitations}
% Although achieving high decoding accuracy and efficiency, \method takes longer training time than the autoregressive Transformer. Thus, the training efficiency is a limitation for \method. The parameters of autoregressive transformer share the training of next-token prediction across positions, thus can be more data efficient than the parallel generation models. Therefore, more efforts are needed to improve the training efficiency of parallel generation models.
%The low training efficiency can also be attributed to no predefined orders, so that the model need to model the orders by itself.

% \paragraph{Broader Impacts}
% The potential negative social impacts of the improved parallel sequence generation models could also be misused to generate fake news or spread disinformation at a larger scale and faster pace. 
%Additionally, if these models are used to automate jobs, it could lead to job displacement and increase economic inequality.

\bibliography{custom}

\clearpage
\appendix

\newpage

\section{Settings of the Synthetic Experiment}
\label{sec:synthetic}
\paragraph{Data}In order to investigate the effect of modalities in learning, we create source-target pairs with known modalities for training. Suppose we have 4 modalities in the synthetic data, the source sequences can be transformed into target sequences with the rule of modality. 
%Table~\ref{tab:synthetic_data} demonstrates the output rules of the 4 target modalities and their data examples. 
We generate source sequences of length 32 with numbers uniformly chosen from $1\sim 5000$, and randomly choose only one modality for each source sequence to create the target sequences. In total, we generate 100000 sequence pairs for training, and 5000 pairs each for validation and test. 

\paragraph{Model Setup}We train AR and NAR models on the synthetic data, with all models built with 4 encoder layers and 4 decoder layers. To learn all the modalities gradually, we train a NAR model with a modality growing process. Specifically, besides the original NAR training loss that learns 4 modalities, we also train the NAR model to learn 2 modalities with the middle decoder layer. To capture only 2 modalities in the middle, we merge modalities by transforming the targets of modality $\RN{1}$ to modality $\RN{2}$ and the targets of modality of $\RN{3}$ to modality $\RN{4}$. With the merged source-target pairs with only 2 modalities, we train the middle decoder layer to fit modality $\RN{2}$ and $\RN{4}$. 

\paragraph{Evaluation}For output quality measurement, we obtain the model outputs $\hat{\bm{y}}$, and compare them with the closest targets $\bm{y}$ in the 4 modalities for all the outputs. The number accuracy is the percentage of $\hat{y}_i=y_i$ and the sequence accuracy is the percentage of $\hat{\bm{y}}=\bm{y}$. To visualize the distribution of different modalities, we use the modality of the closest target as that of the output, and report the proportion for different modalities.
%Finally, we optimize the objective of the middle layer and the last layer jointly. 

\section{Proof for Definition Consistency of the Modality Diffusion Process}
%$q(\bm{y}_t|\bm{y}_0)$ and $q(\bm{y}_{t-1}|\bm{y}_t, \bm{y}_0)$
\label{sec:consistency}
\paragraph{Lemma 1.} With $q(\bm{y}_{t-1}|\bm{y}_t,\bm{y}_0)$ defined in the modality diffusion process $q^\text{MDP}$, for $t\geq 1$, we have:
\begin{equation}
q(\bm{y}_t|\bm{y}_0)=(\gamma_{t} \gP_{t-1} + (1-\gamma_{t}) \gP_{t-1} \odot \bm{y}_0)/Z_{t}
\end{equation}

\noindent \textit{Proof.}
According to the definition in Section~\ref{sec:modality_diffusion}, we have
\begin{equation}
\label{eqn:proof}
q(\bm{y}_t|\bm{y}_0):=(\gamma_{t} \gP_{t-1} + (1-\gamma_{t}) \gP_{t-1} \odot \bm{y}_0)/Z_{t}
\end{equation}
\begin{equation}
\begin{aligned}
q & (\bm{y}_{t-1}|\bm{y}_t,\bm{y}_0) := \gamma_{t-1} \gP_{t-2} +((1-\gamma_{t-1}) \gP_{t-2} \\
 &-\omega_{t}\gP_{t-1}) \odot \bm{y}_0 + \omega_{t}Z_t \bm{y}_0 \odot \bm{y}_t
\end{aligned}
\end{equation}
Since our models predcit the tokens in the sequence independently, the posterior $q(\bm{y}_t|\bm{y}_0)$ can also be decomposed into the form of independent token distributions:
\begin{equation}
\begin{aligned}
&q(\bm{y}_t|\bm{y}_0)=\prod_{i=1}^n q(y^i_t|\bm{y}_0) \quad where \quad \\
& q(y^i_t|\bm{y}_0)=
\begin{cases}
\gamma_t \gP_{t-1}^i/Z_t & y^i_t\neq y^i_0\\
\gP_{t-1}^i /Z_t & y^i_t= y^i_0\\
\end{cases}
\end{aligned}
\end{equation}
Here, $y^i_0$ is the $i$th token of $\bm{y}_0$, and $\gP_{t-1}^i$ is the $i$th token output distribution of $\gP_{t-1}$. In the same way, we can rewrite $q(y^i_{t-1}|\bm{y}_t,\bm{y}_0)$ as:
\begin{equation}
\label{eqn:token_transition}
\begin{cases}
\gamma_{t-1} \gP_{t-2}^i & y_{t-1}^i\neq y^i_0\\
\gP_{t-2}^i-\omega_t (\gP_{t-1}^i -  Z_t \mathds{1}(y^i_0==y^i_{t})) & y_{t-1}^i= y^i_0\\
\end{cases}
\end{equation}
For any $t\geq 2$, we can compute $q(\bm{y}_{t-1}|\bm{y}_0)$ by:
\begin{equation}
    \begin{aligned}
    q(\bm{y}_{t-1}|\bm{y}_0)= & \int_{\bm{y}_t} q(\bm{y}_t|\bm{y}_0)q(\bm{y}_{t-1}|\bm{y}_t, \bm{y}_0)\text{d} \bm{y}_t 
    \end{aligned}
\end{equation}
Thus,
\begin{equation}
\begin{aligned}
q&(y^i_{t-1}=y^i_0|\bm{y}_0)\\
=& (1-q(y^i_t=y^i_0|\bm{y}_0))q(y^i_{t-1}|y^i_t\neq y^i_0,\bm{y}_0) \\
&+ q(y^i_t=y^i_0|\bm{y}_0)q(y^i_{t-1}|y^i_t= y^i_0,\bm{y}_0)\\
= & (1-\gP_{t-1}^i/Z_t)(\gP_{t-2}^i-\omega_t \gP_{t-1}^i) \\
&+(\gP_{t-1}^i/Z_t) (\gP_{t-2}^i-\omega_t (\gP_{t-1}^i-Z_t)) \\
= & \gP_{t-2}^i-\omega_t \gP_{t-1}^i + \gP_{t-1}^i/Z_t \cdot \omega_t Z_t\\
= & \gP_{t-2}^i
\end{aligned}
\end{equation}
And from Eq.\ref{eqn:token_transition}, we can derive:
\begin{equation}
q(y^i_{t-1}|\bm{y}_0)=\gamma_{t-1} \gP_{t-2}^i \quad \text{for} \  y^i_{t-1}\neq y^i_0
\end{equation}
Therefore, after normalization, we have:
\begin{equation}
\begin{aligned}
& q(\bm{y}_{t-1}|\bm{y}_0)=\prod_{i=1}^n q(y^i_{t-1}|\bm{y}_0) \quad where \quad \\
& q(y^i_{t-1}|\bm{y}_0)=
\begin{cases}
\gamma_t \gP_{t-2}^i/Z_{t-1} & y^i_{t-1}\neq y^i_0\\
\gP_{t-2}^i /Z_{t-1} & y^i_{t-1}= y^i_0\\
\end{cases}
\end{aligned}
\end{equation}
Similarly, we can prove that Eq.\ref{eqn:proof} holds for $t\geq 1$.

\section{Additional Results on Machine Translation}
\label{sec:ad_mt_results}
For reference and further comparison, we also report the tokenized BLEU scores and the results based on CTC~\citep{ctc2006graves} rather DAT~\citep{dag2022huang}.

\paragraph{Results with Tokenized BLEU}
As some previous work reports tokenized BLEU scores on the machine translation benchmarks, we also provide the results for tokenized BLEU scores for direct comparison.
 \begin{table*}[h]
\begin{center}
\begin{small}
%\begin{sc}
\resizebox{0.85\linewidth}{!}{
\setlength{\tabcolsep}{1mm}{
\begin{tabular}{ll|c|>{\hspace*{2.5mm}}r@{}r@{}l@{}l|>{\hspace*{2.5mm}}r@{}r@{}l@{}l|>{\hspace*{2.5mm}}r@{}r@{}l@{}l|>{\hspace*{2.5mm}}r@{}r@{}l@{}l|>{\hspace*{1mm}}r@{}r@{}l@{}l|>{\hspace*{2mm}}r@{}l}
\toprule
\multicolumn{2}{l|}{\multirow{2}{*}{\bf Model}} & \bf Iter & \multicolumn{4}{c|}{\bf WMT14} & \multicolumn{4}{c|}{\bf  WMT14} & \multicolumn{4}{c|}{\bf WMT17} & \multicolumn{4}{c|}{\bf WMT17} & \multicolumn{4}{c|}{\bf Average} & \multicolumn{2}{c}{\multirow{2}{*}{\bf  Speedup}}\\
& &  & \multicolumn{4}{c|}{\bf En-De}  & \multicolumn{4}{c|}{\bf De-En} & \multicolumn{4}{c|}{\bf En-Zh} & \multicolumn{4}{c|}{\bf Zh-En}  &  \multicolumn{4}{c|}{\bf Gap} & \\
\midrule
\multirow{3}{*}{AR} & Transformer~{\scriptsize \cite{transformer2017vaswani}} & $M$ & & 27&.6 &  & & 31&.4 & & &  34&.3 & & & 23&.7 & & & \tempneg0&.35 & & 1&.0x \\
& Transformer \textit{base} (Ours) & $M$ & & 27&.81* & & & 31&.96* & & & 34&.65* &  & & 23&.98*  & & & 0& &  & 1&.0x \\
%& Transformer \textit{big} (Ours) & $M$ & & 28&.69* & & & 32&.64* & & & -&- &  & & -&-  & & & -&- &  & 0&.9x \\
% \midrule
% \multirow{3}{*}{Continuous Diffusion} & Diff-LM~{\scriptsize \cite{li2022diffusion}} & 20 & & 16&.55* &  & & 20&.02* &  & & \multicolumn{2}{c}{-} &  & & \multicolumn{2}{c}{-}  &  & & \tempneg11&.60$^\diamondsuit$ &  & 1&.3x\\
% & Diff-LM+MBR decoding & 20 & & 18&.62* &  & & 22&.74* &  & & \multicolumn{2}{c}{-} &  & & \multicolumn{2}{c}{-}  &  & & \tempneg9&.21$^\diamondsuit$ &  & 0&.6x\\
% & Difformer~{\scriptsize \cite{gao2022difformer}} & 20 & & 27&.10$^\ddag$ & & & -&- &  & & \multicolumn{2}{c}{-} &  & & \multicolumn{2}{c}{-} &  & & -&- &  & -&-\\
\midrule
\multirow{4}{*}{Iterative Models} & CMLM~{\scriptsize \cite{cmlm2019ghazvininejad}} & 10 & & 24&.61 &  & & 29&.40 &  & & \multicolumn{2}{c}{-} &  & & \multicolumn{2}{c}{-}  &  & & \tempneg2&.88$^\diamondsuit$ &  & 2&.2x\\
%& SMART~{\scriptsize \cite{smart2020}} & 10 & & 25&.10 & & & 29&.58 &  & & \multicolumn{2}{c}{-} &  & & \multicolumn{2}{c}{-} &  & & \tempneg2&.55$^\diamondsuit$ &  & 2&.2x\\
% DisCo~{\scriptsize\cite{disco2020kasai}} & $\approx$4 & & 25&.64 &  & & \multicolumn{2}{c}{-} &  & & \multicolumn{2}{c}{-} &  & & \multicolumn{2}{c}{-} &  & & \tempneg2&.31$^\diamondsuit$ &  & 3&.5x \\
& Imputer~{\scriptsize\cite{imputermt2020saharia}} & 8 & & 25&.0 & & & \multicolumn{2}{c}{-} &  & & \multicolumn{2}{c}{-} &  & & \multicolumn{2}{c}{-} & & & \tempneg2&.96$^\diamondsuit$ &  & 2&.7x \\
& SUNDAE~{\scriptsize\cite{savinov2022stepunrolled}} & 10 & & 26&.25 &  & & 30&.80 &  & & \multicolumn{2}{c}{-} &  & & \multicolumn{2}{c}{-} &  & & \tempneg1&.36$^\diamondsuit$ &  & 2&.2x \\
& CMLMC~{\scriptsize\cite{cmlmc2021}} & 10 & & 26&.40 &  & & 30&.92 &  & & \multicolumn{2}{c}{-} &  & & \multicolumn{2}{c}{-} &  & & \tempneg1&.23$^\diamondsuit$ &  & 1&.7x \\
& \textsc{DiNoiSer}~{\scriptsize \cite{ye2023dinoiser}} & 20 & & 24&.48 & & & 29&.40 &  & & \multicolumn{2}{c}{-} &  & & \multicolumn{2}{c}{-} &  & & \tempneg2&.68$^\diamondsuit$ &  & -&-\\
\midrule
\multirow{4}{*}{Non-iterative Models} 
%& Vanilla NAR~{\scriptsize\cite{nat2018gu}} & 1 & & 11&.79 &  & & 16&.27 & & & 18&.92 & & & 8&.69 &  & & \tempneg15&.68 &  & 15&.3x \\
& CTC~{\scriptsize\cite{ctc2018libovicky}} & 1 & & 18&.42 &  & & 23&.65 & & & 26&.84 & & & 12&.23 &  & & \tempneg9&.31 &  & 14&.3x \\
% A\bm{x}E~{\scriptsize\cite{axe2020ghazvininejad}} & 1 & & 20&.40 &  & & 24&.90 & & & \multicolumn{2}{c}{-} &  & & \multicolumn{2}{c}{-} & & & \tempneg7&.24$^\diamondsuit$ &  & 14&.2x \\
% & GLAT~{\scriptsize\cite{glat2021qian}} & 1 & & 19&.42 &  & & 26&.51 &  & & 29&.79 & & & 18&.88 &  & & \tempneg5&.95 &  & 15&.3x \\
& OaXE~{\scriptsize\cite{oaxe2021du}} & 1 & & 22&.4 & & & 26&.8 &  & & \multicolumn{2}{c}{-} &  & & \multicolumn{2}{c}{-} &  & & \tempneg5&.28$^\diamondsuit$ &  & 14&.2x \\
& GLAT+CTC~{\scriptsize\cite{glat2021qian}} & 1 & & 25&.02 &  & & 29&.14 &  & & 30&.65 &  & & 19&.92 &  & & \tempneg3&.42  &  & 14&.3x \\
% CTC + DSLP~{\scriptsize\cite{dslp2021huang}} & 1 & & 24&.81 &  & & 28&.33 &  & & \multicolumn{2}{c}{-} &  & & \multicolumn{2}{c}{-} &  & &  \tempneg3&.32$^\diamondsuit$ & & 14&.0x \\
& DAT$^\dag$~{\scriptsize\cite{dag2022huang}} & 1 & & 26&.95* &  & & 30&.73* &  & & 33&.27* &  & & 23&.60* &  & & \tempneg0&.96 &  & 13&.0x \\
%+ BeamSearch & 1 & & 27&.02 & & & 31&.24 & & & 34&.21 &  & & 24&.22 & & & \tempneg0&.43 & & 7&.1x\\
\midrule
\multirow{2}{*}{Ours} 
%& \method~(CTC) \ \  & 1 & & 26&.46 & & & 30&.48 & & & 31&.77 &  & & 20&.87 &  & & \tempneg2&.21 &  & 14&.3x \\
%+ beamsearch\ \  & 1 & -&- & -&- & -&- & -&- & -&- & -&- & -&- & -&- & \tempneg-&- & \tempneg-&- & 14&.6x \\
& \method$^\dag$ \ \  & 1 & & 27&.40 & & & 31&.94 & & & 34&.12 & & & 24&.23 & & & \tempneg0&.18 & & 13&.0x \\
%+ beamsearch\ \  & 1 & & 27&.58 & & & 31&.95 & & & 34&.83 & & & 24&.81 & & & +0&.19 & & 7&.1x \\
& \method$^\dag$ \ \  & 3 & & 28&.57 & & & 32&.08 & & & 35&.09 & & & 24&.86 & & & +0&.55 & & 7&.2x \\
%& \method \textit{big}~(DAT)$^\dag$ \ \  & 3 & & 29&.28 & & & 32&.86 & & & -&- & & & -&- & & & -&- & & 6&.5x \\
\bottomrule
\end{tabular}
}}
%\end{sc}
\end{small}
\end{center}
\caption{The tokenized BLEU scores on WMT14 En$\leftrightarrow$De and WMT17 Zh$\leftrightarrow$En. The average gap is computed against our Transformer implementation. * represents the results are obtained from our re-implementation, and $\diamondsuit$ indicates that the average gap is only computed with available results. For models with $\dag$, we use the Joint-Viterbi decoding proposed by~\citet{shao2022viterbi} for inference. The average gap is computed against the results of our implemented Transformer \textit{base}.}
\label{tab:add_main_result}
\vspace{-0.5em}
\end{table*}

\paragraph{Combination with CTC}
We also conduct experiments for \method with CTC, and the results are presented in Table~\ref{tab:ctc}.
\begin{table}[t]
\centering
\begin{small}
\resizebox{\linewidth}{!}{
\begin{tabular}{l|c|cccc}
\toprule
\multirow{2}{*}{} & Iter& WMT14 & WMT14& WMT17& WMT17\\
& & En-De & De-En & Zh-En & En-Zh\\
\midrule
GLAT+CTC  &1& 24.85 & 28.37 & 30.20 & 17.57\\
\method (CTC) &1& \textbf{25.92} & \textbf{29.98}& \textbf{31.77}& \textbf{20.66}\\
\bottomrule
\end{tabular}  
}
\end{small}
\caption{The results of \method based on CTC}
\label{tab:ctc}
\end{table}

The experimental results show that \method achieves improvements of 1$\sim$3 BLEU scores over GLAT+CTC, demonstrating the effectiveness of \method.
\method can easily combine with various existing methods for parallel generation because \method maintains the decoding process or simply adds more iterations. Specifically, \method keeps the original inference process in the non-iterative setting or forwards the decoder multiple times without intermediate decoding in the iterative setting.

\section{Inference Time Comparison}
\label{sec:infer_speedup}
To provide a more comprehensive comparison of the decoding speedup, as discussed in~\citep{helcl-etal-2022-non}, we measure the inference latency on the WMT14 test set with 1 Nvidia-V100 GPU, and report the inference latency with batch size 1 in Table~\ref{tab:inference_latency}.
Following the setting in~\citet{deepshallow2021kasai}, we also measure the inference latency of a Transformer with 12 encoder layers and 1 decoder layer. 
\begin{table}[!h]
    \centering
    \resizebox{\linewidth}{!}{
    \begin{tabular}{l|cccc}
        \toprule
         &  Transformer & Transformer & \method & \method\\
         & 6-6 & 12-1 & 6-6 (1 iteration) & 6-6 (3 iterations)\\
         \midrule
         Latency & 297.0ms & 108.9ms & 22.8ms & 42.1ms\\
         \bottomrule
    \end{tabular}}
    \caption{The comparison of inference latency. A-B represents the model has A encoder layers and B decoder layers}
    \label{tab:inference_latency}
    \vspace{-1em}
\end{table}

Comparing with the Transformer~(12-1), \method with 3 decoding iterations still has a $2.6\times$ speedup. Although the Transformer with deep encoder and shallow decoder can achieve faster inference, the depth of decoder is important for capability of decoder-only models~\citep{gpt3,openai2023gpt4}. Thus, the comparison for models with 6 decoder layers is also useful.

\section{Hyper-parameters for Experiments}
\label{sec:hyper}
We train our models with fairseq~\citep{fairseq2019ott}\footnote{\url{https://github.com/facebookresearch/fairseq}}.
For machine translation, the dropout is set to 0.1 except En-Ro/Ro-En and the Transformer \emph{big} setting, where the dropout is 0.3. For paraphrasing and image captioning, we use the dropout of 0.3. 
%To balance $L_{t-1}^\text{RGS}$ in the training objective, we set $\lambda_0=1$ and $\sum_{t=1}^{T-1}\lambda_t=0.5$.
Since our modality diffusion process requires the model to capture part of the modality, for the first 100k steps, we train the model to predict the target $\bm{y}_0$ at all steps $t$. For the subsequent training transitions, we train the model with the modality diffusion process. For DAT, the decoding length is set to be 8 times the source length for non-iterative models and 4 times for iterative models.

For the hyper-parameters of the modality diffusion in Eq.~\ref{eqn:transition}, we use a simplified implementation with the interpolation between $\gP_{t-2}$ and $\bm{y}_0$:
\begin{equation}
q(\bm{y}_{t-1}|\bm{y}_t,\bm{y}_0) = \gamma_{t-1} \gP_{t-2} + (1-\gamma_{t-1})\gP_{t-2}\odot \bm{y}_0
\end{equation}
Here, we set $\gamma_{t-1} \in (0,1]$ to be selected from a pre-defined set of values. To achieve a similar effect for interpolating with $\bm{y}_0\odot \bm{y}_t$, we choose $\gamma_{t-1}$ as the maximum value that preserves 90\% of the tokens in $\bm{y}_0\odot \bm{y}_t$. 
With such interpolation, $\bm{y}_{t-1}$ includes most of the target tokens that is correctly predicted in the previous $\bm{y}_t$. Thus, the difference between $\bm{y}_{t}$ and $\bm{y}_0$ gradually reduces as $t$ decreases.

 In terms of residual glancing training, we set the hyper-parameter of glancing schedule $\mu=1/T$. And the hyper-parameter for computing sampling numbers $\alpha$ decrease from 0.5 to 0.1 periodically for iterative \method.

For paraphrase generation, we utilize a Transformer architecture with 4 encoder layers and 4 decoder layers, with a hidden dimension of 256. For image captioning, we use the image features extracted by fast R-CNN~\footnote{\url{https://github.com/peteanderson80/bottom-up-attention}} with a Transformer \emph{base} setting. The maximum training steps are set to 100k for both paraphrase generation and image captioning.
\begin{figure*}[h]
\begin{minipage}{0.32\linewidth}
\centering
\includegraphics[width=\linewidth]{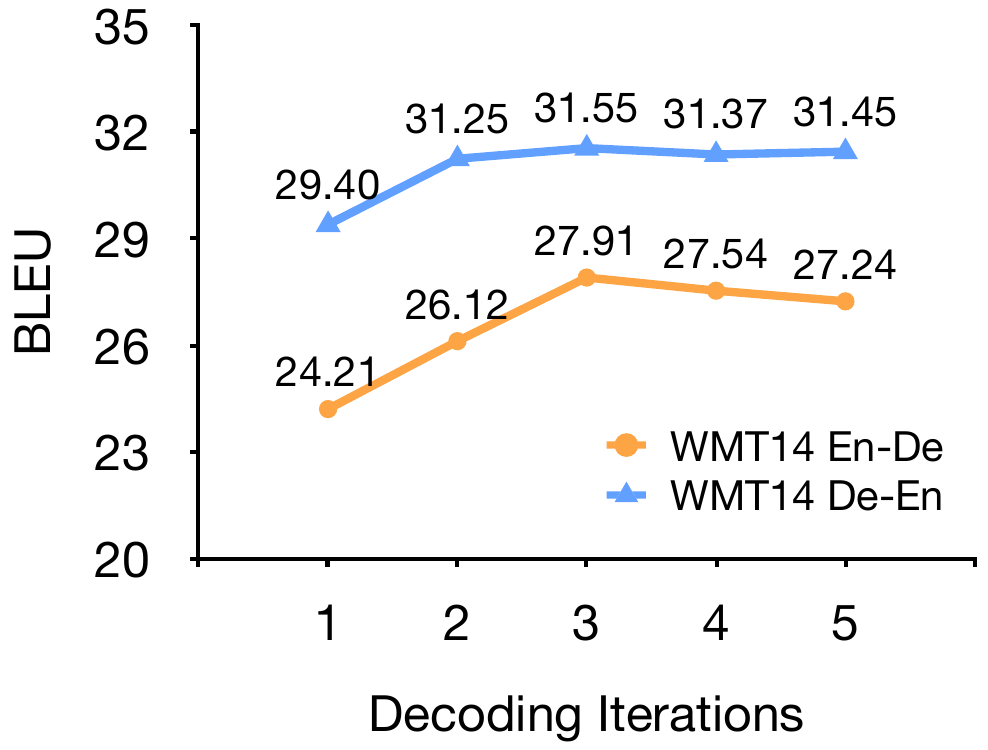}
\caption{The BLEU score curves with iterative decoding.}
\label{fig:iteration}
\end{minipage}\hfill
\begin{minipage}{0.32\linewidth}
\centering
\includegraphics[width=\linewidth]{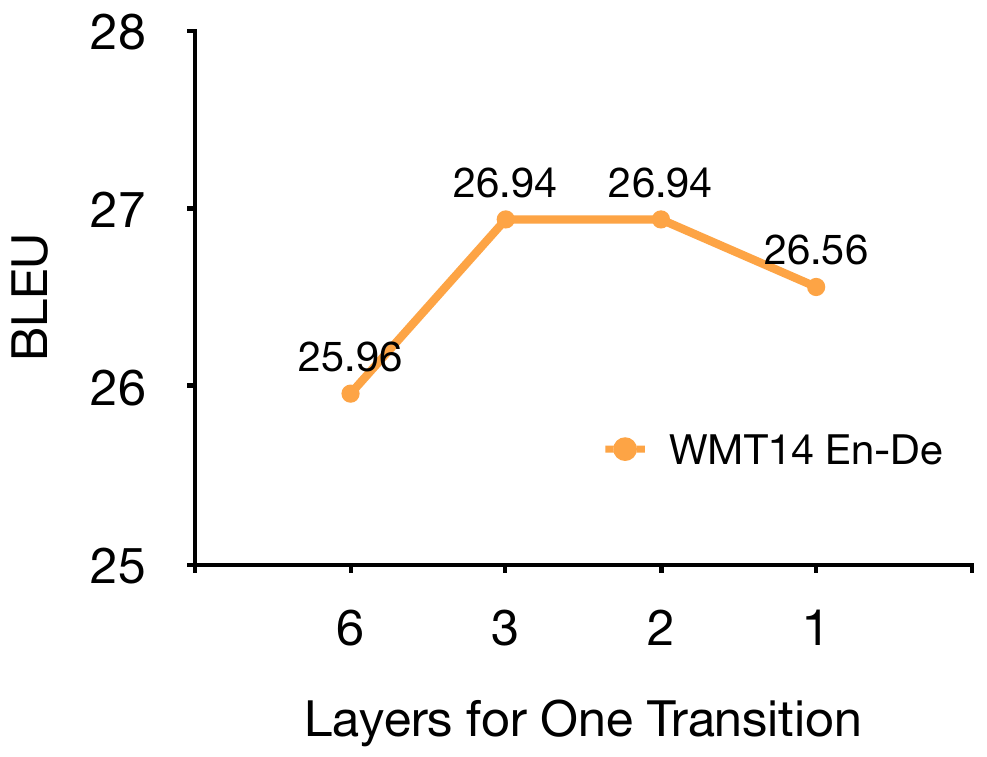}
\caption{Comparison for the number of decoder layers in one denoising diffusion transition.}
\label{fig:layer_nums}
\end{minipage}\hfill
\begin{minipage}{0.32\linewidth}
\centering
\includegraphics[width=\linewidth]{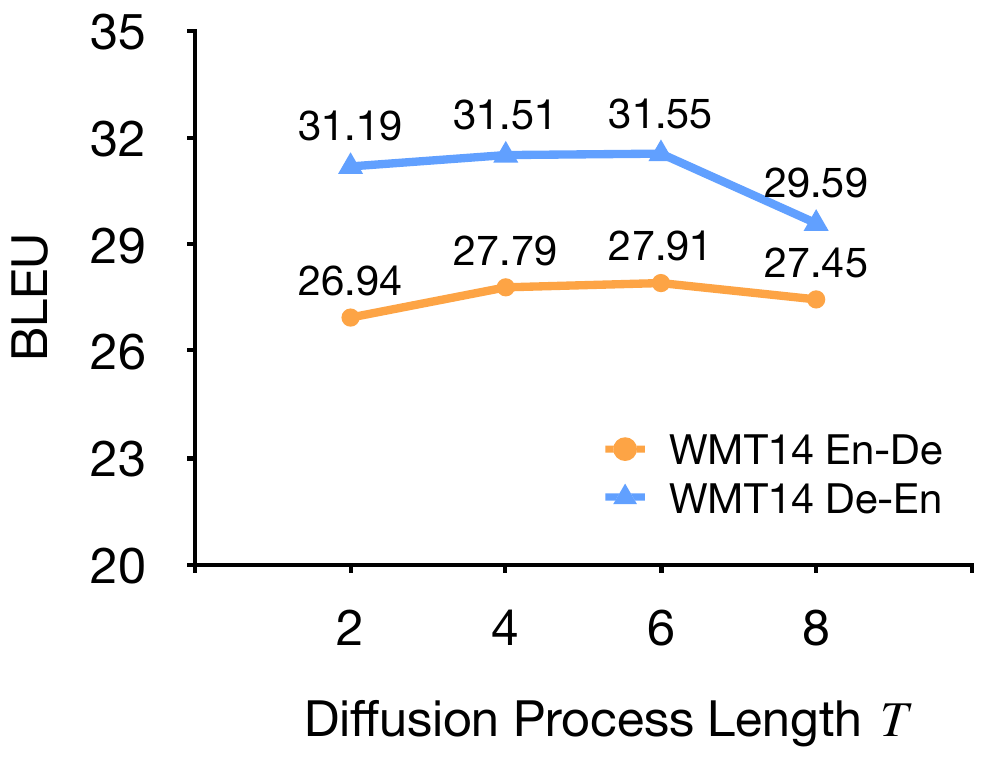}
\caption{The effect of different diffusion process lengths.}
\label{fig:diff_lengths}
\end{minipage}
\end{figure*}

\section{Additional Analysis}
\label{sec:appendix_analysis}
To further study our approach, we analyze the impact of the number of decoding iterations, the number of layers for one reverse transition, and the total diffusion steps in training.
\paragraph{Performance with Iterative Decoding}
The BLEU scores for different decoding iterations are presented in Figure~\ref{fig:iteration}. Here, the evaluated models are trained with generative processes of 3 iterations. Our findings indicate that the BLEU scores increases as the number of iterations increases to 3, and the scores do not increase when the iterations exceeds 3.
Note that the decoding iterations in inference can be different from that in training. Since the model is trained using a generation process of 3 iterations, it also achieves the best scores with 3 decoding iterations, which is consistent with training.
\paragraph{The Number of Decoder Layers for each Transition}
We use 3 decoder layers to model one transition $p_\theta(\bm{y}_{t-1}|\bm{y}_t, \bm{x})$. Thus, one decoding iteration of the 6-layers decoder performs 2 denoising transitions. Without changing the total number of the decoder layers and iterative decoding, we conduct experiments to study the influence of the number of layers for one transition. 
The results in Figure.~\ref{fig:layer_nums} show that the model achieves the best performance when using 2 or 3 layers for one transition. But the scores with only 1 layer for 1 transition decreases, which may caused by insufficient modelling capacity with only 1 layer.
\paragraph{The Effect of Diffusion Process Lengths}
We also investigate the effect of diffusion process steps $T$ and the results are illustrated in Figure~\ref{fig:diff_lengths}. Note that we perform each reverse denoising step with 3 decoder layers, so every iteration corresponds to 2 steps in the diffusion process. We find that the performance grows on WMT14 datasets until $T$ reaches 6, and declines when the $T$ is 8. We think the reason why the performance stops growing with the increasing iterations is that large iterations makes part of the denoising transitions too easy to learn. The easy learning task can lead to capability degradation, as the issues caused by adding small noises in diffusion models for text~\citep{li2022diffusion}.

\section{The chrF and COMET Scores}
\begin{table}[h]
    \centering
    \resizebox{\linewidth}{!}{
    \begin{tabular}{l|cccc}
    \toprule
    & WMT14 En-De & WMT14 De-En& WMT17 En-Zh & WMT17 Zh-En\\
    \midrule
    Transformer & 0.5800  & 0.5832 & 0.3078 & 0.5230\\
    DAT & 0.5621 & 0.5669 & 0.2995 & 0.5090\\
    \method & 0.5790 & 0.5815 & 0.3136 & 0.5310\\
    \bottomrule
    \end{tabular}}
    \caption{The chrF scores on WMT14 En$\leftrightarrow$De and WMT17 En$\leftrightarrow$Zh}
    \label{tab:comet}
    \vspace{-1em}
\end{table}
\begin{table}[h]
    \centering
    \resizebox{\linewidth}{!}{
    \begin{tabular}{l|cccc}
    \toprule
    & WMT14 En-De & WMT14 De-En& WMT17 En-Zh & WMT17 Zh-En\\
    \midrule
    Transformer & 0.8623  & 0.8711 & 0.8599 & 0.8537\\
    DAT & 0.7966 & 0.8470 & 0.8265 & 0.8222\\
    \method & 0.8341 & 0.8600 & 0.8469 & 0.8366\\
    \bottomrule
    \end{tabular}}
    \caption{The COMET scores on WMT14 En$\leftrightarrow$De and WMT17 En$\leftrightarrow$Zh}
    \label{tab:comet}
    \vspace{-1em}
\end{table}
Besides the commonly used BLEU~\citep{bleu2002papineni} metric, we compute the additional chrF~\citep{popovic-2015-chrf} and COMET~\citep{rei-etal-2020-comet} scores for more comprehensive evaluation. We use the wmt22-comet-da for computing the COMET score.

\method achieves better chrF scores on WMT17 En$\leftrightarrow$Zh and comparable scores on WMT14 En$\leftrightarrow$De compared to the Transformer.
And \method still outperforms DAT in terms of COMET but falls behind the Transformer baseline.
The gap in COMET may caused by the distribution mismatch between the NAR outputs and the training data for COMET.

 \section{Data Statistics and Evalution Metrics}
 The statistics of the data we used in experiments are listed as follow: WMT14 En$\leftrightarrow$De~(4.5m), WMT17 En$\leftrightarrow$Zh~(20m), WMT14 En$\leftrightarrow$fr~(35m), WMT16 En$\leftrightarrow$Ro~(0.6m) and WMT13 En$\leftrightarrow$Es(12m). For paraphrase generation and image caption, we use the Quora~(145k) dataset and MS-COCO~(113k) dataset respectively. For MS-COCO, we use the Karpathy split~\citep{karpathy2015deep}.
 
For paraphrase, we report tokenized BLEU and ROUGE-L~\citep{lin2004rouge}\footnote{The script is ROUGE-1.5.5.pl}. For image caption, we also report METEOR~\citep{banerjee2005meteor} and CIDEr~\citep{vedantam2015cider}, and use the official evaluation tools for evaluation\footnote{\url{https://github.com/cocodataset/cocoapi}}.

\section{Connections with Diffusion Models and Iterative NAR}
\label{sec:related}
\paragraph{Diffusion Models for Discrete Data}
%The diffusion models have achieved great success for continuous data~\citep{diffusion2020, song2021denoising}, while the autoregressive models are still in dominance for discrete data. 
%Recently, a series of work begins to model discrete data with diffusion models. 
Previous work studies continuous or discrete process for modelling discrete data with diffusion models. For continuous diffusion processes, a series of work explores adding Gaussian noise in the word embedding space~\citep{li2022diffusion}, converting the discrete data to 0/1 bits~\citep{chen2022analog}, the design space of diffusion models\citep{dieleman2022continuous}, and conditional text generation with continuous diffusion~\citep{gong2022diffuseq}.
%Different from directly utilizing the continuous diffusion process, our method use a process defined in the discrete space.
For discrete diffusion processes, \citet{argmaxflow} study a multinomial diffusion process where each state transits to other states uniformly, while \citet{austin2021discrete} explore more types of state transitions, including masking and increasing the transition frequency for similar states. \citet{he2022diffusionbert} propose a noise schedule based on token information. \citet{zheng2023reparameterized} derive a discrete diffusion framework with route mechanism via reparameterization.

\paragraph{Discussion for comparison with Step-Unrolled Models and Semi-Autoregressive Training}
% The work for non-autoregressive neural machine translation can be divided into non-iterative and iterative methods.
% %The non-iterative methods only forward the neural network once in generation, which keep high inference speed with parallel decoding. 
% Previous non-iterative methods attempt to enhance the generation quality by learning from autoregressive models~\citep{imitatenat2019wei, guo2020fine}, incorporating light sequential decoding~\citep{crf2019sun, dag2022huang}, introducing informative decoder input~\citep{pnat2019bao, flowseq2019ma, song2021alignart}, adaptive training strategies~\citep{glat2021qian, bao2022latent} and training objectives with alignment~\citep{ctc2018libovicky,axe2020ghazvininejad,oaxe2021du}. Besides, models with iterative decoding have also been developed~\citep{iterativerefinement2018lee,cmlm2019ghazvininejad,levenshtein2019gu,dslp2021huang,imputermt2020saharia,savinov2022stepunrolled,cmlmc2021}.
~\citet{savinov2022stepunrolled} introduces a denoising procedure related to diffusion models but different from ours. Specifically, SUNDAE corrupts the target and uses the corrupted sequence as the input for learning denoising. In contrast, \method employs a diffusion process for dividing modalities, and uses the corrupted sequence as the intermediate training target. Besides, SUNDAE heavily relies on multiple decoding iterations, while our method can achieve competitive quality without iterative decoding.

Compared with SMART~\citep{smart2020}, our method learns to predict the intermediate targets rather than replacing decoder inputs. For SMART, the model first generates outputs with several iterations and uses the outputs as the decoding inputs to learn mistake correction. In contrast, DIFFGLAT uses the intermediate targets to capture the multi-modality in data gradually. Although both methods can be used for iterative refinement, they work on input and target, respectively

\end{document}